\documentclass{article}

\usepackage[main, final]{neurips_2026}

\usepackage[utf8]{inputenc} 
\usepackage[T1]{fontenc}    
\usepackage{hyperref}       
\usepackage{url}            
\usepackage{booktabs}       
\usepackage{amsfonts}       
\usepackage{nicefrac}       
\usepackage{microtype}      
\usepackage{xcolor}         
\usepackage{amsmath}

\usepackage{graphicx}

\title{H2G: Hierarchy-Aware Hyperbolic Grouping \\for 3D Scenes}

\author{
  Byungha Ko$^{1,2}$\thanks{Equal contribution.} \quad Youngmin Lee$^{1,2\ast}$ \quad Dong Hwan Kim$^{1,2}$ \\[0.6em]
  $^1$ Department of Computer Science and Engineering, Korea University \\
  $^2$ Intelligence and Interaction Research Center, Korea Institute of Science and Technology \\[0.4em]
  \texttt{\{rhqo123, aidi0724, gregorykim\}@kist.re.kr}
}

\begin{document}

\maketitle

\begin{abstract}
Hierarchical 3D grouping aims to recover scene groups across multiple granularities, from fine object parts to complete objects, without relying on semantic labels or a fixed vocabulary. The main challenge is to transform 2D foundation-model cues into coherent hierarchy supervision and embed that hierarchy in a 3D representation. We propose H2G, a hyperbolic affinity field for hierarchical 3D grouping. Our method derives semantically organized tree supervision by interpreting foundation-model affinities through Dasgupta's objective for similarity-based hierarchical clustering. This supervision is distilled into a single Lorentz hyperbolic feature field, whose geometry is well suited for tree-like branching structures. A hierarchy-aware objective aligns the field with fine-level assignments, coarse object structure, compact feature clusters, and LCA (Lowest Common Ancestor) ordering. This formulation represents multiple grouping levels in one feature space, enabling semantic hierarchical grouping grounded in 2D foundation-model knowledge. Code is available at \url{https://github.com/Rhqo/H2G}.

\end{abstract}

\section{Introduction}

3D scene grouping is inherently hierarchical: the same point may belong to an object part, a whole object, or a larger object group depending on the desired granularity. Recent methods lift category-agnostic 2D masks, such as SAM\cite{sam} proposals, into NeRF-style\cite{nerf} 3D fields for multi-view consistent grouping. The central challenge is not only to lift masks into 3D, but to organize overlapping and conflicting 2D proposals into a coherent hierarchy.

However, how these lifted masks should be organized into a hierarchy remains under-specified. GARField\cite{garfield} resolves grouping ambiguity with a scale-conditioned affinity field, but this formulation ties hierarchy to depth-derived physical scale, which can be unreliable for large objects, partially observed objects, and amorphous background regions. It also requires repeated rendering over query scales. OmniSeg3D\cite{omniseg3d} and Ultrametric Feature Fields\cite{uff} avoid explicit scale queries through a unified feature-field representation, but the resulting hierarchies are still largely driven by mask-derived relations such as overlap, containment, or pixel-level correspondences among SAM-generated\cite{sam} masks. These relations provide useful boundary structure, but they do not necessarily align with semantic hierarchy.

Therefore, we propose H2G, a hyperbolic affinity field for hierarchical 3D grouping. Our method converts SAM\cite{sam} proposals into non-overlapping regions, computes DINO\cite{dinov3} descriptors for them, and constructs semantic affinity graphs over the resulting regions. Since their affinities are symmetric pairwise similarities rather than ordered hierarchy labels, we reconstruct them into tree supervision using a top-down construction motivated by Dasgupta's objective\cite{dasgupta} for similarity-based hierarchical clustering.

The resulting 2D hierarchy is distilled into a single 3D grouping field in Lorentz hyperbolic space. The negative curvature of hyperbolic space provides a natural bias for representing branching structures with many descendants, while our hierarchy-aware objective combines leaf separation, root separation, cluster compactness, and LCA (Lowest Common Ancestor) ordering. This enables scale-free hierarchical 3D grouping from semantic relations, without depth-based scale queries or repeated scale-conditioned rendering.

Our contributions are threefold:
\begin{itemize}
    \item We introduce a semantic 2D hierarchy construction that converts SAM\cite{sam} proposals and DINO\cite{dinov3} region affinities into tree supervision using a top-down procedure motivated by Dasgupta's hierarchical clustering\cite{dasgupta} objective.
    \item We formulate hierarchical 3D grouping as a scale-free Lorentz hyperbolic feature field, representing fine-to-coarse groups in a single embedding space without scale-conditioned rendering.
    \item We design a hierarchy-aware learning objective that combines leaf separation, root-level grouping, cluster compactness, and LCA-order preservation to align the 3D feature geometry with the constructed hierarchy.
\end{itemize}

The rest of this paper is organized as follows. Section~\ref{sec:related_work} reviews neural feature fields, hierarchical 3D grouping, and hyperbolic representation learning. Section~\ref{sec:method} describes the construction of 2D hierarchy supervision and its distillation into a Lorentz hyperbolic feature field. Section~\ref{sec:experiments} presents quantitative and qualitative evaluations, followed by the conclusion.

\section{Related work}
\label{sec:related_work}
\paragraph{Neural Feature Fields for 3D Scene Understanding}

Neural radiance fields (NeRF)\cite{nerf} have been extended beyond RGB and density to store semantic, instance, language, and foundation-model features in 3D\cite{nesf, kobayashi2022decomposing, liu2023weakly, semantic-nerf, panoptic-lift, panoptic-nerf, distilled-ff}. LERF\cite{lerf}, OpenScene\cite{openscene}, and OpenNeRF\cite{opennerf} align neural fields or 3D points with vision-language features for open-vocabulary scene understanding, while Contrastive Lift\cite{contrastive-lift} shows that 2D segmentation cues can be lifted into 3D for instance segmentation. These works establish neural feature fields as a useful substrate for storing foundation-model signals in 3D. However, they typically support language queries, category retrieval, or flat instance labels, rather than class-agnostic hierarchical grouping. H2G instead uses feature fields to store hierarchy-aware grouping cues.

\paragraph{Hierarchical 3D Grouping from 2D Masks}

Closest to our setting are methods that lift multi-granular 2D masks, such as SAM\cite{sam} proposals, into view-consistent 3D representations\cite{garfield, omniseg3d, uff, sa3d}. GARField\cite{garfield} resolves grouping ambiguity with a scale-conditioned affinity field, Ultrametric Feature Fields\cite{uff} reveal hierarchy by thresholding feature distances, and OmniSeg3D\cite{omniseg3d} distills hierarchical 2D mask representations into a single 3D feature field. These methods show the importance of hierarchy, but differ in its source. Scale-conditioned inference requires repeated rendering across query scales, while mask overlap or containment may reflect proposal geometry more than semantic organization. H2G differs by reconstructing the 2D hierarchy from DINO\cite{dinov3} semantic affinities and embedding all grouping levels in a single Lorentz hyperbolic field.

\paragraph{Hyperbolic Representation Learning for Hierarchical Structure}

Hyperbolic representation learning\cite{poincare, lorentz, hyperbolic-image, hyperbolic-survey} is well suited to tree-like data because hyperbolic volume grows exponentially with radius, allowing branching structures to be embedded with lower distortion than in Euclidean space. This geometry has been used for symbolic hierarchies, hierarchical clustering\cite{hyphc, cross-hyphc}, and vision-language representations\cite{meru, atmg}. MERU\cite{meru} models visual-semantic specificity in hyperbolic image-text space, HypHC\cite{hyphc} uses hyperbolic geometry for hierarchical clustering and LCA recovery, and OpenHype\cite{weijler2025openhype} introduces hyperbolic embeddings into open-vocabulary radiance fields. These works motivate hyperbolic geometry as an inductive bias for hierarchy, but do not directly address dense class-agnostic 3D grouping. H2G applies this inductive bias to dense 3D grouping by learning ray-level Lorentz features with angular and LCA-order objectives.

\section{Method}
\label{sec:method}
\begin{figure}[t]
  \centering
  \includegraphics[width=\textwidth]{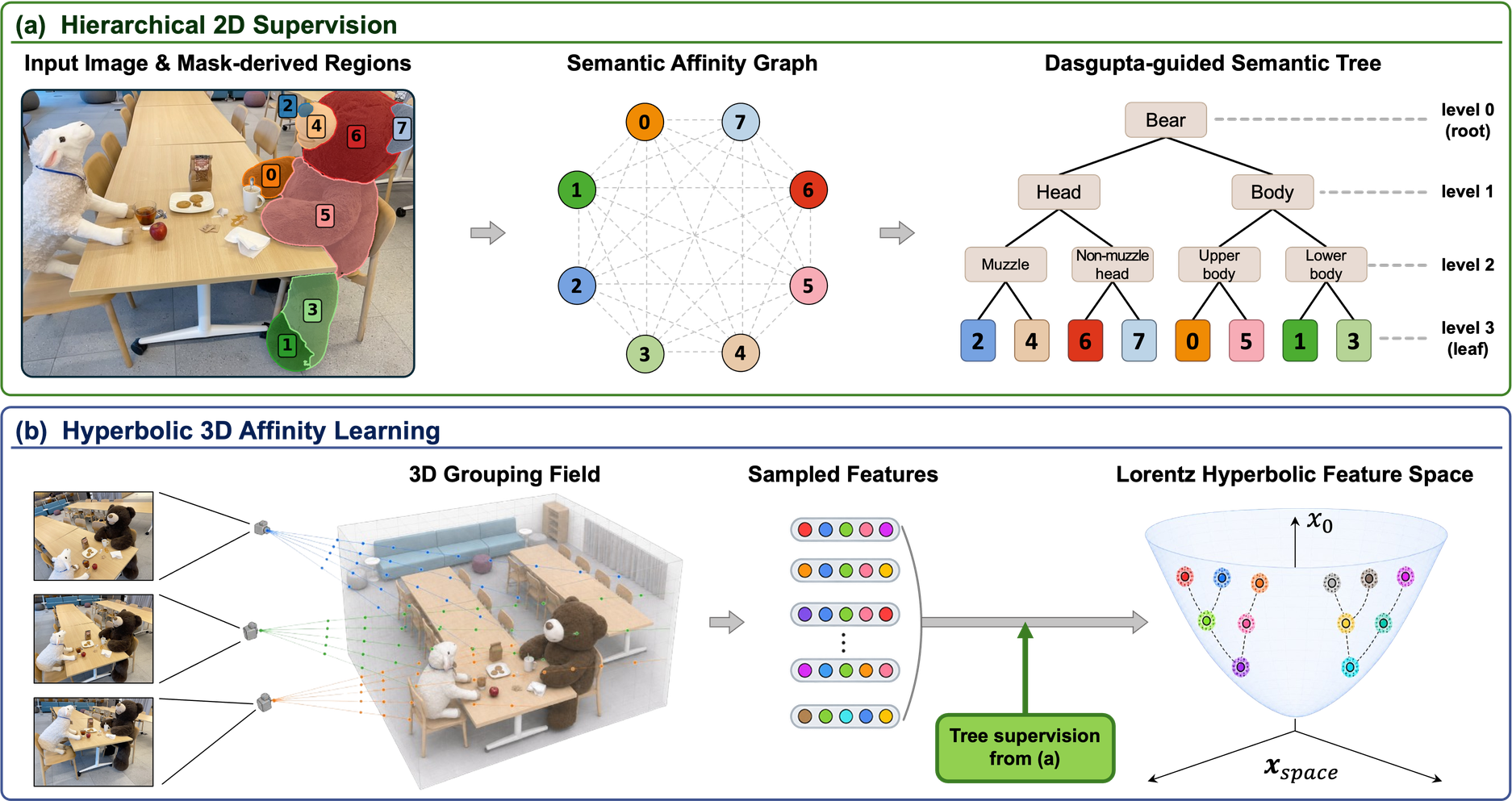}
  \caption{Overview of H2G. (a) SAM leaf masks are connected by DINO affinities and organized into a Dasgupta-guided semantic tree. (b) The tree supervises the rendered ray features of a NeRF-style grouping field in a Lorentz feature space. Text labels in the tree are illustrative; H2G does not use semantic class labels for supervision.}
  \label{fig:overview}
\end{figure}

Figure~\ref{fig:overview} summarizes the H2G pipeline. H2G first constructs image-wise hierarchy supervision from SAM\cite{sam} and DINO\cite{dinov3} cues, then distills it into a Lorentz hyperbolic feature field. The 2D stage converts overlapping SAM\cite{sam} proposals into non-overlapping leaf masks, organizes them into semantic trees using DINO\cite{dinov3} affinities, and provides leaf, root, and LCA relations for 3D learning. This design follows prior observations that SAM masks typically nest as whole, part, and subpart\cite{sam} and that the similarity of self-supervised features decreases from fine to coarse granularity\cite{rossetti2024hier}.

\subsection{Constructing 2D Hierarchy Supervision}
\label{app:hierarchy_supervision}
Let \(\{I^n\}_{n=1}^{N}\) denote the training images, where \(n\) is the image index. Given \(I^n\), we first collect SAM\cite{sam} mask proposals and assign each proposal to its closest containing parent, defined as the smallest larger proposal whose containment ratio exceeds a threshold; proposals without parents become roots. Proposals sharing the same root are processed together, and their overlaps are resolved into fine non-overlapping leaf masks with unique ancestor paths. We denote one such leaf \(a\) by \(m_a^n\).

Each leaf \(m_a^n\) receives a DINO\cite{dinov3} descriptor \(\mathbf z_a^n\) by average-pooling the DINO feature map \(\Phi^n\) over its covered patches and applying \(\ell_2\) normalization:

$$
\bar{\mathbf z}_a^n
=
\frac{1}{|\Omega_a^n|}
\sum_{\mathbf{u}\in\Omega_a^n}
\Phi^n(\mathbf{u}),
\quad
\mathbf z_a^n
=
\frac{\bar{\mathbf z}_a^n}{\|\bar{\mathbf z}_a^n\|_2}.
$$

where \(\Omega_a^n\) is the set of covered DINO\cite{dinov3} patch indices and \(\mathbf u\) indexes a DINO patch location. For each root-induced group indexed by \(i\), we define an affinity graph \(\mathcal G_i^n=(\mathcal V_{\mathrm{leaf},i}^n,\mathcal E_i^n,\mathbf W_i^n)\), where \(\mathcal V_{\mathrm{leaf},i}^n\) contains the induced leaf vertices and \(\mathcal E_i^n\) fully connects them. The affinity matrix entry for an edge \((u,v)\in\mathcal E_i^n\) for each group \(i\) is
\((\mathbf W_i^n)_{uv}=W_{uv}^n=\max(0,(\mathbf z_{u}^n)^\top\mathbf z_{v}^n)\). The graph is symmetric, whereas the desired supervision is an ordered tree. Dasgupta's objective\cite{dasgupta} for similarity-based hierarchical clustering provides the guiding principle for this conversion. For a tree \(\mathcal T\) and pairwise similarities \(\mathbf W\), the objective assigns cost

$$
\mathcal C(\mathcal T;\mathbf W)
=
\sum_{a<b}
W_{ab}
\left|
\operatorname{Leaf}
\left(
\operatorname{LCA}_{\mathcal T}(a,b)
\right)
\right|.
$$

Here \(\operatorname{LCA}_{\mathcal T}(a,b)\) is the lowest common ancestor of leaves \(a\) and \(b\), and \(\operatorname{Leaf}(v)\) is the set of leaf descendants under node \(v\). This objective favors trees where high-affinity leaves meet in small subtrees. Dasgupta's top-down approximation recursively applies an \(\alpha\)-approximation algorithm for sparsest cut\cite{dasgupta}, but solving such cuts on our dense DINO\cite{dinov3} graphs is expensive. We therefore use recursive spectral bisection\cite{fiedler1975property} as a practical surrogate. This surrogate gives comparable costs with substantially lower preprocessing time; we report this comparison in Appendix~\ref{app:spectral_bisection}.

Starting from each affinity graph, we recursively split the current leaf set to build a tree. At each split, we compute the Fiedler vector of the normalized graph Laplacian of the induced DINO\cite{dinov3} affinity subgraph and divide leaves by its sign. Each split creates a virtual internal vertex whose children are the two resulting subtrees. Recursion continues until singleton leaves, with optional flattening of adjacent internal nodes for multi-way relations. Applying this process to all root-induced groups in image \(I^n\) yields a 2D hierarchy forest \(\mathcal T^n=\{\mathcal T_i^n\}_{i=1}^{R_n}\), i.e., a collection of trees, one for each group. 

Each tree \(\mathcal T_i^n=(\mathcal V_{\mathcal T,i}^n,\mathcal E_{\mathcal T,i}^n)\) contains leaf vertices and virtual internal vertices, with \(\mathcal V_{\mathcal T,i}^n=\mathcal V_{\mathrm{leaf},i}^n\cup\mathcal V_{\mathrm{int},i}^n\). Thus, each labeled pixel or ray has a leaf label and a leaf-to-root ancestor chain
\(\mathcal A(y)=\left(y,\operatorname{par}(y),\ldots,\operatorname{root}(y)\right).\)

\begin{figure}[t]
  \centering
  \includegraphics[width=\textwidth]{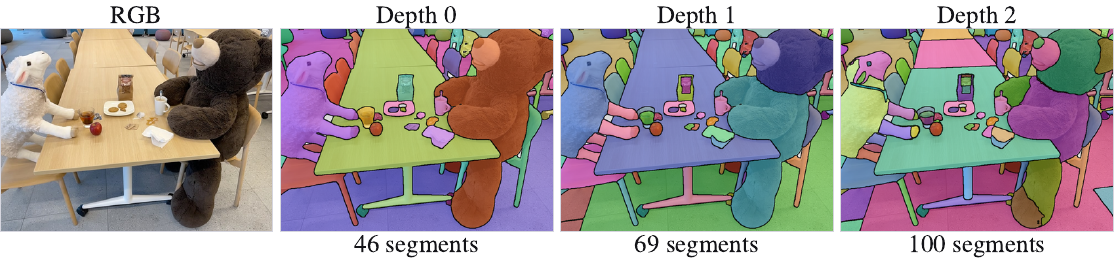}
  \caption{Example of 2D hierarchy supervision. Shallower leaves keep their own segments, and a segment keeps its color until it is split.}
  \label{fig:hierarchy_teatime}
\end{figure}

Figure~\ref{fig:hierarchy_teatime} shows an example of the resulting 2D hierarchy supervision, with each tree of \(\mathcal T^n\) cut at depth \(d\) from its root. The forest encodes object--part relations, with the bear at depth 0 as the parent of its head and body at depth 1. The decomposition follows semantics rather than spatial connectivity, as the four disconnected legs of the sheep form one segment at depth 1.

The DINO\cite{dinov3} graph is used only to construct the tree; 3D learning uses the resulting leaf labels, root labels, ancestor relations, and LCA relations as hierarchy supervision.

\subsection{Hyperbolic 3D Affinity Learning}
\paragraph{Hyperbolic Affinity Field}
The 2D hierarchy provides supervision for pixels or rays in each training view, but the goal is not to learn independent 2D segmentations for each view. Instead, H2G learns a multi-view consistent 3D grouping field. To this end, we use a NeRF-style\cite{nerf} 3D feature field. The RGB/density branch provides scene geometry and volume-rendering weights, while the grouping branch predicts features at the same 3D sample locations. These 3D sample features are rendered along each ray to obtain a ray-level grouping feature aligned with the 2D hierarchy.

We use the \(D\)-dimensional Lorentz model \(\mathbb H_c^D\subset\mathbb R^{D+1}\) with curvature parameter \(c>0\). For \(\mathbf{x}=(x_0,\mathbf{x}_{\mathrm{space}})\) and \(\mathbf{y}=(y_0,\mathbf{y}_{\mathrm{space}})\), the Lorentz inner product is \(\langle\mathbf{x},\mathbf{y}\rangle_L=-x_0y_0+\mathbf{x}_{\mathrm{space}}^\top\mathbf{y}_{\mathrm{space}}\), and \(d_{\mathbb H}\) denotes the corresponding geodesic distance\cite{lorentz}. A Euclidean network output \(\mathbf u\in\mathbb R^D\) is projected to the Lorentz hyperboloid by

$$
\Pi_c(\mathbf u)
=
\left(
\sqrt{\frac{1}{c}+\|\mathbf u\|_2^2},
\mathbf u
\right)
\in\mathbb H_c^D .
$$

Concretely, for sample points \(\mathbf x_{r,k}\) on ray \(r\), a hashgrid encoder\cite{mueller2022instant} \(E_\psi\) predicts sample features, and a projection head \(G_\psi\) maps the rendered feature to a Euclidean tangent parameter before hyperbolic projection. We aggregate the sample features over a high-weight sample set \(\mathcal K_r\). Let \(\tilde w_{r,k}\) denote normalized rendering weights detached from the gradient path. The ray-level feature and its hyperbolic projection are

$$
\bar{\mathbf h}_r
=
\sum_{k\in\mathcal K_r}
\tilde w_{r,k}E_\psi(\mathbf x_{r,k}),
\quad
\mathbf s_r
=
\Pi_c\left(G_\psi(\bar{\mathbf h}_r)\right)
\in\mathbb H_c^D .
$$

Here \(\mathbf s_r\) is the hyperbolic grouping feature for ray \(r\), and it serves as the basic unit for the hierarchy-aware losses below.

\paragraph{Hierarchical Prototypes}
Direct pixel- or ray-wise contrastive learning over all pairs is costly and creates many redundant comparisons among rays from the same mask. Inspired by ProtoNCE~\cite{protonce}, H2G treats each observed leaf mask as a prototype target: ray features assigned to the same leaf are aggregated into a single mask-level representative for contrastive supervision. This reduces the number of comparison targets and stabilizes learning by averaging noisy ray features within each leaf.

Let \(\mathcal B\) be a mini-batch of rays sampled from image \(I^n\). Each ray \(r\in\mathcal B\) has a leaf label \(y_r\) from the non-overlapping leaf partition, and the observed leaf set is \(\mathcal V_{\mathcal B}^{\mathrm{leaf}}=\{y_r\mid r\in\mathcal B\}\). For batches spanning multiple images, the same definitions are applied per image. We instantiate prototypes only for leaves in \(\mathcal V_{\mathcal B}^{\mathrm{leaf}}\). Virtual internal vertices are not prototype targets; they provide root and LCA metadata through the corresponding forest. To respect the Lorentz geometry, we aggregate ray features using the Klein-coordinate-based Einstein midpoint\cite{klein}:

$$
\kappa(\mathbf{x})=\frac{\mathbf{x}_{\mathrm{space}}}{x_0},
\quad
\operatorname{E_{mid}}(\mathcal X)
=
\kappa^{-1}
\left(
\frac{\sum_{\mathbf{x}\in\mathcal X}\gamma(\mathbf{x})\kappa(\mathbf{x})}
{\sum_{\mathbf{x}\in\mathcal X}\gamma(\mathbf{x})}
\right).
$$
Here \(\gamma(\mathbf{x})=(1-\|\kappa(\mathbf{x})\|_2^2)^{-1/2}\), and \(\kappa^{-1}\) denotes the inverse map from Klein coordinates to the Lorentz hyperboloid\cite{klein}.

Using this operator, the prototype for each observed leaf label \(\ell\in\mathcal V_{\mathcal B}^{\mathrm{leaf}}\) is computed from its assigned ray features:
\(\mathbf{p}_\ell=\operatorname{E_{mid}}\left(\{\mathbf{s}_r\mid y_r=\ell,\ r\in\mathcal B\}\right).\) The resulting prototype set is \(\mathcal P_{\mathcal B}^{\mathrm{leaf}}=\{\mathbf p_\ell\mid \ell\in\mathcal V_{\mathcal B}^{\mathrm{leaf}}\}\).

\paragraph{Angular Separation Loss}
Following the hyperbolic angular formulation of ATMG~\cite{atmg}, H2G uses angular classification at two levels. The leaf-level loss aligns each ray feature with its assigned leaf prototype for fine discrimination, while the root-level loss aligns each leaf prototype with its root centroid for coarse group separation. Both losses use the same hyperbolic exterior angle. For the hyperbolic triangle formed by the origin \(O\), a reference point \(\mathbf p\), and a query point \(\mathbf s\), we use the angle at \(\mathbf p\), computed in the Lorentz model as

$$
\theta(\mathbf{p},\mathbf{s})
=
\cos^{-1}
\left(
\frac{s_0+p_0c\langle\mathbf{p},\mathbf{s}\rangle_L }
{\|\mathbf{p}_{\mathrm{space}}\|_2\sqrt{(c\langle\mathbf{p},\mathbf{s}\rangle_L)^2-1}}
\right).
$$

A smaller \(\theta(\mathbf p,\mathbf s)\) indicates stronger directional alignment. We use angle rather than hyperbolic distance so that semantic identity is organized mainly by direction, while radial position remains available for compactness and hierarchy ordering.

For leaf-level supervision, each ray feature \(\mathbf s_r\) is classified against the leaf prototypes observed in the current batch:

$$
\mathcal L_{\mathrm{leaf}}
=
-\frac{1}{|\mathcal B|}
\sum_{r\in\mathcal B}
\log
\frac{
\exp\left(-\theta(\mathbf p_{y_r},\mathbf s_r)/\tau_{\mathrm{leaf}}\right)
}{
\sum_{v\in\mathcal V_{\mathcal B}^{\mathrm{leaf}}}
\exp\left(-\theta(\mathbf p_v,\mathbf s_r)/\tau_{\mathrm{leaf}}\right)
}.
$$

This loss separates leaves in angular space while aligning rays with their assigned leaf prototype.

Leaf-level separation alone does not guarantee that different root-level groups are well separated. The purpose of the root-level loss is not to classify individual rays again, but to make leaf-level representatives under the same root share a coarse group direction. Therefore, the classified queries are leaf prototypes rather than ray features. Let \(\mathcal R_{\mathcal B}=\{\operatorname{root}(\ell)\mid \ell\in\mathcal V_{\mathcal B}^{\mathrm{leaf}}\}\) be the set of roots observed in the current batch. For each root \(\rho\in\mathcal R_{\mathcal B}\), we compute a root centroid as the Einstein midpoint of all ray features in the batch whose leaves descend from \(\rho\):

$$
\mathbf{q}_\rho
=
\operatorname{E_{mid}}
\left(
\{\mathbf{s}_r
\mid
\operatorname{root}(y_r)=\rho,\ r\in\mathcal B\}
\right).
$$

Each leaf prototype \(\mathbf p_\ell\) then uses its root \(\operatorname{root}(\ell)\) as the target class and is classified against the root centroids \(\{\mathbf q_\rho\}\):

$$
\mathcal L_{\mathrm{root}}
=
-\frac{1}{|\mathcal V_{\mathcal B}^{\mathrm{leaf}}|}
\sum_{\ell\in\mathcal V_{\mathcal B}^{\mathrm{leaf}}}
\log
\frac{
\exp\left(-\theta(\mathbf q_{\operatorname{root}(\ell)},\mathbf p_\ell)/\tau_{\mathrm{root}}\right)
}{
\sum_{\rho\in\mathcal R_{\mathcal B}}
\exp\left(-\theta(\mathbf q_\rho,\mathbf p_\ell)/\tau_{\mathrm{root}}\right)
}.
$$

This term aligns leaf prototypes with their corresponding root centroids, encouraging leaves under the same root to share a coarse direction.

\paragraph{Compactness Loss}
Angle-based supervision separates semantic directions, but it does not control within-cluster spread. The compactness loss contracts only ray features that lie outside a margin around their assigned prototype.

For each ray feature \(\mathbf s_r\), we penalize the excess hyperbolic distance to its assigned leaf prototype \(\mathbf p_{y_r}\):

$$
\mathcal L_{\mathrm{comp}}
=
\frac{1}{|\mathcal B|}
\sum_{r\in\mathcal B}
\left(
ReLU(
d_{\mathbb H}
\left(
\mathbf{s}_r,
\operatorname{sg}(\mathbf{p}_{y_r})
\right)
-\epsilon)
\right)^2.
$$

Here \(\operatorname{sg}(\cdot)\) denotes the stop-gradient operation, and \(\epsilon\) is the compactness margin. Thus, rays within the margin receive no compactness gradient, while rays outside the margin are pulled toward the current prototype.

\paragraph{LCA Ordering Loss}
Angular losses separate leaves and roots, but they do not encode the merging order inside each tree. We adopt the LCA-order surrogate from HypHC~\cite{hyphc} and apply it to tree-derived leaf prototypes in our 3D grouping field. LCA order therefore ranks prototype pairs by tree-derived ancestor depth. Let \(\mathbf{k}_i=\kappa(\mathbf{p}_i)\) and \(\mathbf{k}_j=\kappa(\mathbf{p}_j)\) be the Klein coordinates of two leaf prototypes. Since Klein geodesics are straight segments, the continuous LCA surrogate is approximated by the point on the segment closest to the origin:

$$
t_{ij}
=
\operatorname{clip}_{[0,1]}
\left(
-\frac{\mathbf{k}_i^\top(\mathbf{k}_j-\mathbf{k}_i)}
{\|\mathbf{k}_j-\mathbf{k}_i\|_2^2}
\right),
\quad
\widehat{\mathbf{k}}_{ij}
=
\mathbf{k}_i+t_{ij}(\mathbf{k}_j-\mathbf{k}_i).
$$

We use the hyperbolic distance from the origin to this point as a pairwise LCA-depth surrogate:

$$
d_o(i,j)
=
\frac{1}{\sqrt c}
\cosh^{-1}
\left(
\frac{1}{\sqrt{1-\|\widehat{\mathbf{k}}_{ij}\|_2^2}}
\right).
$$

Triplets \((i,j,k)\in\mathcal P_{\mathrm{LCA}}\) are sampled so that \(i\) and \(j\) come from the same parent, while \(k\) comes from a different sibling branch under the same grandparent. The desired relation is

$$
d_o(i,j) > d_o(i,k),
\quad
d_o(i,j) > d_o(j,k).
$$

This relation is optimized with a softmax ranking objective:

$$
\mathcal L_{\mathrm{LCA}}
=
-\frac{1}{|\mathcal P_{\mathrm{LCA}}|}
\sum_{(i,j,k)\in\mathcal P_{\mathrm{LCA}}}\!\!\!\!\!\!\!\!\!
\log
\frac{
\exp(d_o(i,j)/\tau_{\mathrm{LCA}})
}{
\exp(d_o(i,j)/\tau_{\mathrm{LCA}})
+\exp(d_o(i,k)/\tau_{\mathrm{LCA}})
+\exp(d_o(j,k)/\tau_{\mathrm{LCA}})
}.
$$

This loss enforces only relative LCA-depth ordering, preserving the merge order of the 2D hierarchy without regressing absolute pairwise distances.

\paragraph{Overall Objective}
The final training objective is the weighted sum of the four hierarchy losses and a max-norm regularizer:
$$
\mathcal L
=
\lambda_{\mathrm{leaf}}\mathcal L_{\mathrm{leaf}}
+\lambda_{\mathrm{root}}\mathcal L_{\mathrm{root}}
+\lambda_{\mathrm{comp}}\mathcal L_{\mathrm{comp}}
+\lambda_{\mathrm{LCA}}\mathcal L_{\mathrm{LCA}}
+\lambda_{\mathrm{norm}}\mathcal L_{\mathrm{norm}} .
$$

Rather than using a single contrastive objective for all hierarchy relations, H2G decomposes supervision into four complementary constraints: leaf-level angular separation, root-level angular grouping, within-leaf compactness, and LCA-order preservation. The max-norm regularizer only prevents Lorentz features from drifting too far from the origin and does not define hierarchy relations.

Through this decomposition, H2G distills the image-wise hierarchy obtained from SAM\cite{sam} and DINO\cite{dinov3} into a 3D hyperbolic affinity field. Each ray therefore receives a hyperbolic feature that reflects not only a leaf-level grouping label, but also root-level group structure and tree ordering.

H2G uses a NeRF\cite{nerf} framework and training setup that closely follows GARField~\cite{garfield}, while replacing scale-conditioned Euclidean grouping features with Lorentz features trained by our hierarchy-aware objective. Additional architecture and training details are provided in Appendix~\ref{app:implement_details}.

\section{Experiments}
\label{sec:experiments}
We evaluate whether H2G can identify and recover 3D groups across fine-to-coarse levels of detail within a single learned feature space.
\subsection{Dataset}

We evaluate on the GARField\cite{garfield} scenes with hierarchical annotations. For 3D Completeness, each scene provides three evaluation views with query points corresponding to the same 3D location and ground-truth masks at two or three hierarchy levels. For group recall, each scene provides ground-truth hierarchical groups for best-match proposal evaluation.

\paragraph{Benchmark protocol}
Following the GARField 3D completeness protocol\cite{garfield}, we use query-conditioned mask retrieval across hierarchy levels. H2G renders each view once and obtains candidate masks by thresholding a tangent-space affinity map computed from the query feature, without scale input or per-level rendering. Additional evaluation details are provided in Appendix~\ref{app:evaluation_details}.

For fair comparison, we compare H2G against baselines reproduced under the same evaluation protocol, denoted by $^{*}$. Published SAM\cite{sam} and GARField\cite{garfield} results are included only as references. We report OmniSeg3D\cite{omniseg3d} in Appendix~\ref{app:full_completeness} because reproduced results are unavailable for one evaluation scene; the main table focuses on baselines with complete scene coverage.

\subsection{3D Completeness}
We report 3D completeness under two thresholding settings: view-wise selection, which chooses the best threshold for each view and level, and level-wise selection, which shares one threshold per hierarchy level across views. Each level is averaged over the scenes that annotate it, and the Mean averages the resulting level values. All averages over scenes and levels in this paper follow this convention, detailed in Appendix~\ref{app:aggregation}.

\begin{table}[t]
\centering
\caption{Average 3D Completeness under view-wise (V) and level-wise (L) threshold selection. $^{*}$ denotes methods evaluated under our reproduced setting.}
\label{tab:completeness_results}
{
\small
\setlength{\tabcolsep}{4.5pt}
\begin{tabular}{lcccccccc}
\toprule
& \multicolumn{2}{c}{Fine} & \multicolumn{2}{c}{Medium} & \multicolumn{2}{c}{Coarse} & \multicolumn{2}{c}{Mean} \\
\cmidrule(lr){2-3}\cmidrule(lr){4-5}\cmidrule(lr){6-7}\cmidrule(lr){8-9}
Model & V & L & V & L & V & L & V & L \\
\midrule
SAM\cite{sam}           & 0.594 & -- & 0.741 & -- & 0.887 & -- & 0.741 & -- \\
GARField\cite{garfield} & 0.846 & -- & 0.843 & -- & 0.933 & -- & 0.874 & -- \\
\midrule
SAM$^{*}$      & 0.557 & -- & 0.742 & -- & \textbf{0.901} & -- & 0.733 & -- \\
GARField$^{*}$ & 0.824 & 0.823 & 0.607 & 0.587 & 0.815 & 0.784 & 0.749 & 0.731 \\
Ours           & \textbf{0.848} & \textbf{0.840} & \textbf{0.813} & \textbf{0.809} & 0.844 & \textbf{0.828} & \textbf{0.835} & \textbf{0.826} \\
\bottomrule
\end{tabular}
}
\end{table}

Table~\ref{tab:completeness_results} shows that H2G achieves the best reproduced mean in both settings, improving over GARField$^{*}$ from 0.749 to 0.835 with view-wise thresholds and from 0.731 to 0.826 with level-wise thresholds. The gains are largest at the medium level, where semantic grouping is less directly tied to physical scale, supporting the role of semantic hierarchy supervision. The full scene-wise breakdown is reported in Appendix~\ref{app:full_completeness}; on \texttt{living\_room}, GARField$^{*}$ remains slightly better in the view-wise mean.

We also introduce Cross-view Threshold Retention (CTR), the ratio between level-wise and view-wise 3D Completeness averaged over scenes. CTR measures how much performance is retained when one threshold is shared across views for each hierarchy level.

\begin{table}[t]
\centering
\caption{Cross-view Threshold Retention (CTR).
Each value is computed as level-wise 3D Completeness divided by view-wise 3D Completeness for each scene, then averaged across scenes.}
\label{tab:ctr_results}
\begin{tabular}{lcccc}
\toprule
Method & Fine & Medium & Coarse & Mean \\
\midrule
GARField$^{*}$ & \textbf{0.998} & 0.967 & 0.959 & 0.975 \\
Ours           & 0.990 & \textbf{0.995} & \textbf{0.978} & \textbf{0.988} \\
\bottomrule
\end{tabular}
\end{table}

Table~\ref{tab:ctr_results} shows that both methods retain most of their view-wise performance, while H2G has higher overall retention. Since both methods use a shared 3D representation, CTR should not be interpreted as multi-view consistency alone; rather, it reflects how consistently hierarchy levels are placed along each method's inference variable. The higher mean CTR suggests that H2G calibrates medium and coarse levels more consistently in feature affinity, without relying on an explicit scale query.

\subsection{Group Recall}
We follow the GARField group-recall protocol\cite{garfield} to evaluate whether each method generates useful group candidates. GARField\cite{garfield} accumulates candidates from scale-conditioned renderings, while H2G generates candidates from a single rendered feature field. We report native recall, fixed-budget recall, and budget-curve AUC; details are in Appendix~\ref{app:evaluation_details}.

\begin{table}[t]
\centering
\caption{Group recall. Native uses all generated candidates. Budgeted rows sample at most \(K\) candidates without ground-truth information and average over 100 seeds. AUC values are normalized to the range of each metric.}
\label{tab:group_recall_budget}
{\scriptsize
\setlength{\tabcolsep}{3.2pt}
\begin{tabular}{lcccccc}
\toprule
& \multicolumn{3}{c}{GARField$^{*}$} & \multicolumn{3}{c}{Ours} \\
\cmidrule(lr){2-4}\cmidrule(lr){5-7}
Budget & mIoU & \(R@0.50\) & \(R@0.75\) & mIoU & \(R@0.50\) & \(R@0.75\) \\
\midrule
Candidates & \multicolumn{3}{c}{2010.8} & \multicolumn{3}{c}{314.8} \\
Native & 0.758 & 0.892 & 0.619 & \textbf{0.780} & \textbf{0.896} & \textbf{0.746} \\
\midrule
\(K=150\)  & 0.368 & 0.373 & 0.210 & \textbf{0.541} & \textbf{0.580} & \textbf{0.413} \\
\(K=500\)  & 0.613 & 0.692 & 0.439 & \textbf{0.780} & \textbf{0.896} & \textbf{0.746} \\
\(K=1000\) & 0.710 & 0.831 & 0.563 & \textbf{0.780} & \textbf{0.896} & \textbf{0.746} \\
\(K=2000\) & 0.747 & 0.884 & 0.612 & \textbf{0.780} & \textbf{0.896} & \textbf{0.746} \\
\midrule
\(\mathrm{AUC}_{500}\)  & 0.427 & 0.457 & 0.274 & \textbf{0.596} & \textbf{0.662} & \textbf{0.521} \\
\(\mathrm{AUC}_{2000}\) & 0.641 & 0.741 & 0.494 & \textbf{0.734} & \textbf{0.837} & \textbf{0.689} \\
\bottomrule
\end{tabular}
}
\end{table}

Table~\ref{tab:group_recall_budget} shows that H2G produces a substantially smaller proposal pool while achieving higher native-pool mIoU and recall. H2G also outperforms GARField$^{*}$ at every evaluated proposal budget across mIoU, \(R@0.50\), and \(R@0.75\), and obtains higher budget-curve AUC. These results indicate that the learned hierarchy does not merely recover groups through dense proposal enumeration; instead, it concentrates useful hierarchical groups into a compact candidate pool.

\subsection{Effect of Hyperbolic Geometry}
\label{sec:geometry}
To isolate the effect of the geometry, we train a Euclidean counterpart of H2G whose grouping features lie on the unit hypersphere, the standard Euclidean embedding in contrastive learning~\cite{simclr, hypersphere}, with each hyperbolic operation replaced by its Euclidean counterpart. We compare both geometries at dimensions \(D=256\) and \(D=16\) under the same 2D supervision, as detailed in Appendix~\ref{app:geometry}.

\begin{table}[t]
\centering
\caption{Scene-wise 3D Completeness (view-wise) of unit-norm Euclidean and Lorentz features with identical supervision and objective. Scene rows report per-scene means, and $\Delta$ denotes the difference of Lorentz relative to Euclidean.}
\label{tab:geo_completeness}
{
\small
\setlength{\tabcolsep}{5pt}
\begin{tabular}{lcccccc}
\toprule
& \multicolumn{3}{c}{\(D=16\)} & \multicolumn{3}{c}{\(D=256\)} \\
\cmidrule(lr){2-4}\cmidrule(lr){5-7}
Scene & Euclidean & Lorentz & \(\Delta\) & Euclidean & Lorentz & \(\Delta\) \\
\midrule
bouquet      & 0.676 & \textbf{0.742} & +0.066 & 0.749 & \textbf{0.750} & +0.001 \\
teatime      & 0.916 & \textbf{0.953} & +0.037 & 0.951 & \textbf{0.952} & +0.001 \\
ramen        & 0.662 & \textbf{0.807} & +0.145 & 0.631 & \textbf{0.859} & +0.228 \\
living\_room & 0.729 & \textbf{0.815} & +0.086 & 0.776 & \textbf{0.824} & +0.048 \\
\midrule
Mean         & 0.734 & \textbf{0.817} & +0.083 & 0.761 & \textbf{0.835} & +0.074 \\
\bottomrule
\end{tabular}
}
\end{table}

Table~\ref{tab:geo_completeness} shows that Lorentz features achieve a higher mean at both dimensions, but the gain depends on the scene and the dimension. At \(D=256\), the two geometries are nearly identical on \texttt{bouquet} and \texttt{teatime}, while \texttt{ramen} and \texttt{living\_room} clearly favor the Lorentz field, so with enough dimensions a Euclidean field can also represent the hierarchy on some scenes. At \(D=16\), the Lorentz field leads on every scene. On three of the four scenes, it loses little when \(D\) is reduced from 256 to 16, whereas the Euclidean field drops noticeably, consistent with prior work showing that hyperbolic space can represent hierarchies with few dimensions~\cite{poincare}.

\subsection{Ablation study}

\begin{table}[t]
\centering
\caption{Ablation study on the proposed loss components.}
\label{tab:ablation}
\begin{tabular}{lcccc}
\toprule
Variant & Fine & Medium & Coarse & Mean \\
\midrule
\textbf{Full} & 0.848 & 0.813 & \textbf{0.844} & \textbf{0.835} \\
w/o angle     & 0.631 & 0.774 & 0.830 & 0.745 \\
w/o coarse    & \textbf{0.862} & \textbf{0.878} & 0.585 & 0.775 \\
w/o compact   & 0.843 & 0.771 & 0.782 & 0.799 \\
w/o LCA       & 0.847 & 0.806 & 0.839 & 0.831 \\
\bottomrule
\end{tabular}
\end{table}

Table~\ref{tab:ablation} shows that angular separation is the most important component for fine-level grouping, while the root-level term controls the trade-off between local and coarse grouping. Removing the coarse term improves Fine and Medium scores, but sharply reduces Coarse completeness, leading to a lower overall mean. This suggests that coarse supervision intentionally sacrifices some local thresholdability to preserve larger root-level groups. Compactness also improves the thresholdability of groups; LCA-order has a smaller effect on this metric but regularizes the relative merge order of leaf prototypes.

\begin{figure}[!t]
  \centering
  \includegraphics[width=\textwidth]{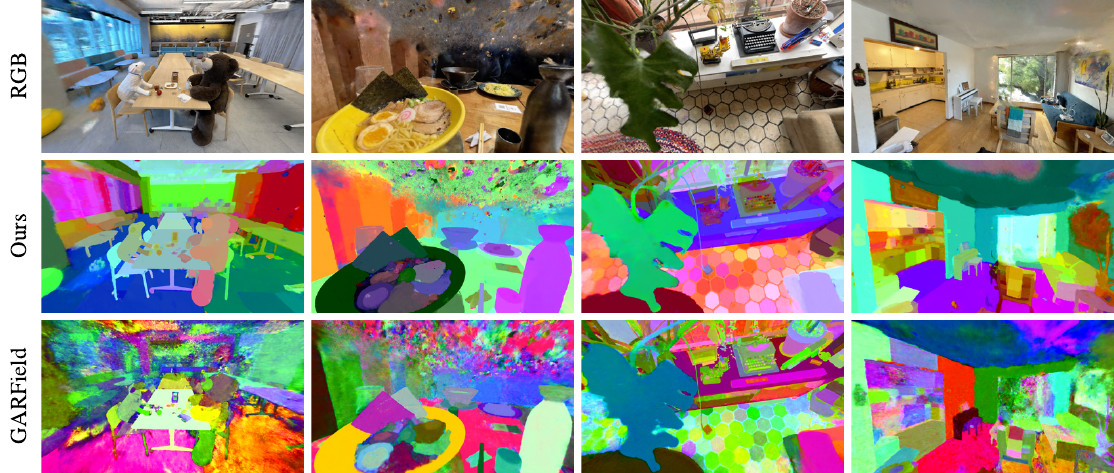}
  \caption{PCA (Principal component analysis) visualization of rendered grouping features. From left to right, the scenes are \texttt{teatime}, \texttt{ramen}, \texttt{living\_room}, and \texttt{bouquet}. GARField is shown at scale 0 for comparison.}
  \label{fig:qual_pca}
\end{figure}

\subsection{Qualitative results}

\begin{figure}[!t]
  \centering
  \includegraphics[width=\textwidth]{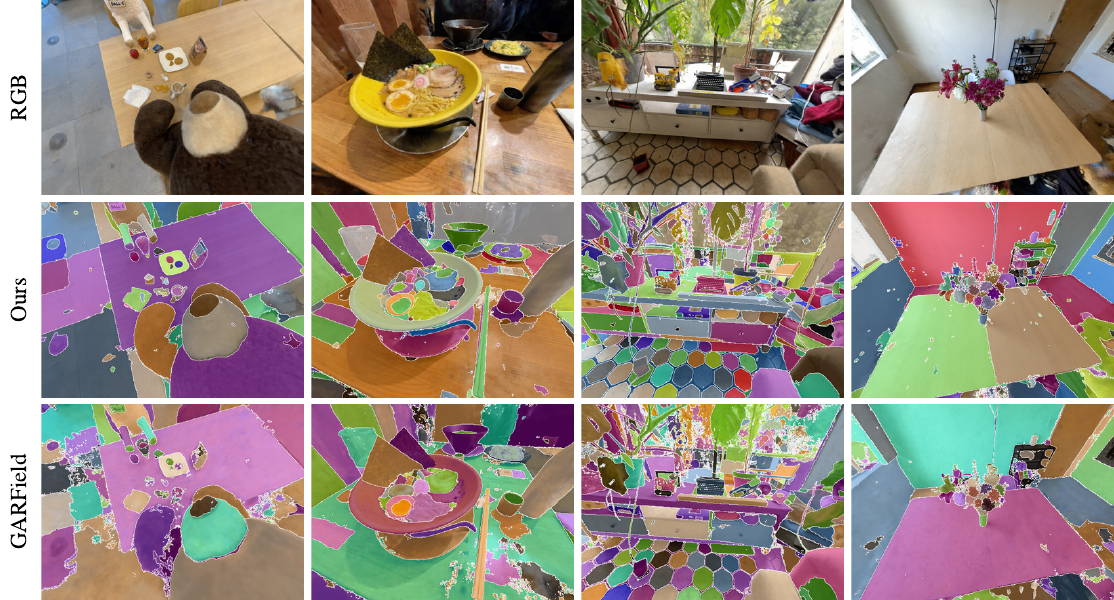}
  \caption{HDBSCAN clustering of rendered grouping features. From left to right, the scenes are \texttt{teatime}, \texttt{ramen}, \texttt{living\_room}, and \texttt{bouquet}. GARField clusters are visualized from its recursive scale-sweep refinement, while H2G clusters are obtained from a single rendered feature field.}
  \label{fig:qual_hdbscan}
\end{figure}

Figures~\ref{fig:qual_pca} and~\ref{fig:qual_hdbscan} visualize properties that are not fully captured by IoU-based scores. Compared with GARField\cite{garfield}, H2G produces clearer feature regions on large surfaces and more structured fine-grained clusters around object parts in cluttered or visually ambiguous regions. These qualitative differences suggest that part- and object-level groups form coherent neighborhoods in a single feature space, even when the final oracle IoU does not increase for every annotated group.

\section{Limitations}
H2G depends on the quality of its 2D supervision. Noisy SAM\cite{sam} proposals or unreliable DINO\cite{dinov3} affinities can produce imperfect hierarchy trees, especially for small, occluded, or visually ambiguous regions. Since trees are constructed per image, some cross-view inconsistencies may remain after 3D distillation. Moreover, recursive spectral bisection is a practical approximation to hierarchy construction, and final mask extraction still requires thresholding or clustering in the learned feature space. Finally, our evaluation is limited to the four scenes of the GARField benchmark\cite{garfield}, which to our knowledge is the only public 3D dataset with both cross-view hierarchical query masks and hierarchical group annotations.

\section{Conclusion}
We presented H2G, a hierarchy-aware hyperbolic grouping field for 3D scenes. H2G reconstructs tree supervision from SAM\cite{sam} proposals and DINO\cite{dinov3} affinities, and distills it into a single Lorentz feature field through leaf, root, compactness, and LCA-order objectives. On the GARField 3D Completeness benchmark\cite{garfield}, H2G improves over reproduced baselines under both view-wise and level-wise threshold selection, especially at medium granularity. These results show that semantic hierarchy supervision and hyperbolic geometry provide an effective basis for hierarchical 3D grouping without scale-conditioned rendering.

\section*{Acknowledgments}
This work was supported by the KIST Institutional Program (Project No. 26E0210, 26E0213).   

{
\small
\bibliographystyle{unsrt}
\bibliography{Docs/references}
}


\appendix

\section{Implementation details}
\label{app:implement_details}
We train H2G in two stages. First, we optimize the RGB radiance field for 30k iterations. We then freeze the RGB field, proposal networks, and camera optimizer, and train only the semantic grouping field for 20k iterations with hierarchy supervision enabled from the beginning.

For 2D hierarchy preprocessing, SAM\cite{sam} masks are generated in automatic mode and sorted by area. We form top-down proposal groups using a containment threshold of 0.8, convert overlapping masks into non-overlapping leaf masks by assigning pixels to the smallest covering mask, and apply morphological closing and connected-component filtering. DINO\cite{dinov3} descriptors are mean-pooled over the resulting leaf masks, and neighboring internal nodes with a small height gap can be flattened to represent multi-way relations.

The grouping feature dimension is 256. During semantic training, we sample 8192 rays from one image per iteration and optimize the semantic branch with Adam. The initial learning rate is \(10^{-3}\), the final learning rate is \(10^{-4}\), warmup lasts 1000 steps, weight decay is \(10^{-6}\), and gradients are clipped with maximum norm 1.0. Unless otherwise stated, we use \(\lambda_{\mathrm{leaf}}=1.0\), \(\lambda_{\mathrm{root}}=2.0\), \(\lambda_{\mathrm{comp}}=1.0\), \(\lambda_{\mathrm{LCA}}=0.05\), and \(\lambda_{\mathrm{norm}}=1.0\), with temperatures \(\tau_{\mathrm{leaf}}=0.22\), \(\tau_{\mathrm{root}}=0.20\), and \(\tau_{\mathrm{LCA}}=0.5\), compactness margin \(\epsilon=0.1\), and curvature \(c=1.0\). All experiments were run on a local workstation equipped with NVIDIA RTX 3090 GPUs, each with 24GB VRAM, except the inference timings in Appendix~\ref{app:inference}, which were measured on an NVIDIA RTX 4090. Generating the 2D hierarchy supervision takes approximately 30 minutes per scene, and hierarchy-supervised grouping-field training takes approximately 20 minutes per scene.

\subsection{Reproduced Baselines}
\label{app:reproducibility}

For the reproduced baselines, we used the same GARField\cite{garfield} evaluation scenes and query annotations, and matched the available SAM ViT-H checkpoint and mask-generation settings as closely as possible. We observed differences from the published results, which can arise from checkpoint availability, implementation details, environment differences, and the sensitivity of scale-conditioned grouping to ambiguous same-scale groups. The GARField$^{*}$ rows in Tables~\ref{tab:completeness_results} and~\ref{tab:completeness_scene_results} come from a single run at the default seed (42), and we did not search over seeds. Below, we re-evaluate both GARField and H2G over five runs with seeds 0--4.

\paragraph{Seed sensitivity}
Neither the GARField paper~\cite{garfield} nor its official implementation specifies a seed, and the evaluation protocol document distributed with the code notes that grouping conflicts within a single scale can resolve either way and that, in this case, ``perhaps changing the random seed could be helpful.'' We reran GARField five times, varying the seed of the official implementation. The spread differed sharply across cells (scene \(\times\) level): \texttt{teatime} Coarse ranged over 0.752--0.964 and \texttt{ramen} Medium over 0.542--0.904, while some cells stayed within \(\pm\)0.007.

\paragraph{Matched-seed comparison}
Running H2G over the same seeds, under the same SAM masks and the same protocol, gives the 3D Completeness in Table~\ref{tab:seed_completeness}. H2G still leads in mean, although by a smaller margin than in the single-seed Table~\ref{tab:completeness_results}. The two methods differ in how evenly they cover the levels. GARField is strongest at Coarse but weakest at Medium, where its run-to-run standard deviation is also an order of magnitude larger than that of H2G. H2G's three levels stay close to one another, so even its weakest level remains well above GARField's Medium.

\begin{table}[h]
\centering
\caption{3D Completeness (view-wise) over seeds 0--4, aggregated as in Table~\ref{tab:completeness_results}. Each entry is the mean \(\pm\) standard deviation over the five seeds.}
\label{tab:seed_completeness}
{
\small
\begin{tabular}{lcccc}
\toprule
Method & Fine & Medium & Coarse & Mean \\
\midrule
GARField$^{*}$ & 0.826 \(\pm\) 0.009 & 0.669 \(\pm\) 0.075 & \textbf{0.900} \(\pm\) 0.020 & 0.798 \(\pm\) 0.029 \\
Ours           & \textbf{0.852} \(\pm\) 0.007 & \textbf{0.801} \(\pm\) 0.004 & 0.832 \(\pm\) 0.014 & \textbf{0.829} \(\pm\) 0.004 \\
\bottomrule
\end{tabular}
}
\end{table}

\paragraph{Dependence on SAM}
Both methods take SAM\cite{sam} masks as their input, so what SAM produces carries directly into their 3D results. For these runs, we used the checkpoint GARField specifies (\texttt{sam\_vit\_h\_4b8939.pth}) with the implementation that checkpoint belongs to. The pipeline nevertheless turned out to be sensitive to how SAM is run. The original release (fb) and the HuggingFace port (hf) of that same checkpoint are indistinguishable in 2D under our evaluation, yet they lead GARField to different 3D results (Table~\ref{tab:sam_fb_hf}). This does not suggest that we used a different SAM from the authors; it indicates how far GARField's 3D output can move on details that leave 2D quality untouched.

\begin{table}[h]
\centering
\caption{Sensitivity to the SAM implementation. Masks come from the original release (fb) or the HuggingFace port (hf) of the same SAM ViT-H checkpoint. 2D evaluates the SAM masks alone, as for SAM$^{*}$ in Table~\ref{tab:completeness_results}; 3D is the view-wise 3D Completeness of GARField$^{*}$ averaged over seeds 0--4. Scene rows are scene means, and the Mean row follows Appendix~\ref{app:aggregation}.}
\label{tab:sam_fb_hf}
{
\small
\begin{tabular}{lcc}
\toprule
Scene & SAM masks, 2D (fb / hf) & GARField$^{*}$, 3D (fb / hf) \\
\midrule
teatime      & 0.845 / 0.844 & 0.902 / 0.911 \\
bouquet      & 0.514 / 0.514 & 0.748 / 0.673 \\
ramen        & 0.799 / 0.801 & 0.789 / 0.721 \\
living\_room & 0.809 / 0.809 & 0.832 / 0.813 \\
\midrule
Mean         & 0.733 / 0.733 & 0.798 / 0.754 \\
\bottomrule
\end{tabular}
}
\end{table}

Like the seed, other settings that neither the GARField paper nor its code specifies may have influenced the published results. We therefore mark every reproduced result with $^{*}$, and our claims are made against these reproduced baselines rather than against the reference results.

\section{Evaluation Details}
\label{app:evaluation_details}

\subsection{3D Completeness Protocol}
For each scene, we evaluate H2G using the annotated camera poses, query pixels, and COCO-format\cite{coco} hierarchy masks provided by the GARField 3D Completeness benchmark\cite{garfield}. Each evaluation view contains a query pixel corresponding to the same 3D point, and each hierarchy level provides a ground-truth mask that contains the query. H2G renders the RGB image and grouping feature field once for each evaluation camera; unlike scale-conditioned methods, the renderer does not receive a separate scale or hierarchy-level input.

For the reported H2G scores, we use tangent-space query affinity. Let \(f(p)\) be the rendered Lorentz feature at pixel \(p\), and let \(q\) be the query pixel. Features are mapped to the tangent space at the origin with the Lorentz logarithmic map, and the query-conditioned score is computed as

$$
z(p)=\log_0(f(p);c), \quad
\mathrm{score}_q(p)=\exp\left(-\|z(p)-z(q)\|_2^2\right).
$$

A candidate mask for threshold \(t\) is \(M_q(t)=\{p\in I\mid \mathrm{score}_q(p)\geq t\}\). In the view-wise setting, we sweep 1000 linearly spaced thresholds in \([0.0,0.99]\) independently for each evaluation view and annotated hierarchy level, then select the threshold that maximizes IoU with the corresponding ground-truth mask. The per-level score is the mean IoU over the three evaluation views. In the level-wise setting, we use the same tangent affinity maps but select one threshold per hierarchy level by maximizing the total IoU across all evaluation views, then evaluate every view at that level with the shared threshold. The reported mean averages the available Fine, Medium, and Coarse levels for each scene.

\paragraph{Aggregation over scenes and levels}
\label{app:aggregation}
The evaluation scenes do not annotate the same set of hierarchy levels. \texttt{bouquet}, \texttt{ramen}, and \texttt{living\_room} provide Fine, Medium, and Coarse masks, whereas \texttt{teatime} provides two levels, which we assign to Fine and Coarse. Each level is averaged over the scenes that annotate it, i.e., Fine and Coarse over four scenes and Medium over three, and the Mean averages the resulting level values. The Mean is therefore not the average of the scene-wise Means. This applies to both view-wise and level-wise scores in Tables~\ref{tab:completeness_results} and~\ref{tab:completeness_scene_results}, while a scene-wise Mean in Table~\ref{tab:completeness_scene_results} averages the levels annotated in that scene. CTR in Table~\ref{tab:ctr_results} follows the same order: the ratio is computed per scene and level, averaged per level over the scenes that annotate it, and the Mean averages the level values. For the published SAM\cite{sam} and GARField\cite{garfield} results, we take the scene-wise values and recompute every average under this convention. Group recall has no level annotations, so the aggregate values in Table~\ref{tab:group_recall_budget} are averages over the four scenes.

\begin{table}[!t]
\centering
\caption{Scene-wise 3D Completeness under view-wise (V) and level-wise (L) threshold selection. $^{*}$ denotes methods evaluated under our reproduced setting. OmniSeg3D$^{*}$ is not included in the mean rows because its \texttt{living\_room} result is unavailable.}
\label{tab:completeness_scene_results}
{
\scriptsize
\setlength{\tabcolsep}{2.5pt}
\begin{tabular}{llcccccccc}
\toprule
& & \multicolumn{2}{c}{Fine} & \multicolumn{2}{c}{Medium} & \multicolumn{2}{c}{Coarse} & \multicolumn{2}{c}{Mean} \\
\cmidrule(lr){3-4}\cmidrule(lr){5-6}\cmidrule(lr){7-8}\cmidrule(lr){9-10}
Scene & Model & V & L & V & L & V & L & V & L \\
\midrule
teatime & SAM\cite{sam}           & 0.816 & -- & -- & -- & 0.973 & -- & 0.895 & -- \\
        & GARField\cite{garfield}       & 0.927 & -- & -- & -- & 0.979 & -- & 0.953 & -- \\
\cmidrule(lr){2-10}
        & SAM$^{*}$      & 0.714 & -- & -- & -- & \textbf{0.976} & -- & 0.845 & -- \\
        & GARField$^{*}$ & 0.893 & 0.893 & -- & -- & 0.754 & 0.633 & 0.824 & 0.763 \\
        & OmniSeg3D$^{*}$& 0.306 & 0.281 & -- & -- & 0.249 & 0.192 & 0.278 & 0.237 \\
        & Ours           & \textbf{0.931} & \textbf{0.928} & -- & -- & 0.973 & \textbf{0.972} & \textbf{0.952} & \textbf{0.950} \\
\midrule
bouquet & SAM\cite{sam}           & 0.174 & -- & 0.735 & -- & 0.761 & -- & 0.557 & -- \\
        & GARField\cite{garfield}       & 0.760 & -- & 0.816 & -- & 0.854 & -- & 0.810 & -- \\
\cmidrule(lr){2-10}
        & SAM$^{*}$      & 0.126 & -- & 0.584 & -- & \textbf{0.831} & -- & 0.514 & -- \\
        & GARField$^{*}$ & \textbf{0.785} & \textbf{0.785 }& 0.617 & 0.563 & 0.632 & \textbf{0.632} & 0.678 & 0.660 \\
        & OmniSeg3D$^{*}$& 0.388 & 0.388 & 0.702 & 0.697 & 0.267 & 0.254 & 0.453 & 0.447 \\
        & Ours           & 0.753 & 0.748 & \textbf{0.862} & \textbf{0.858} & 0.635 & 0.601 & \textbf{0.750} & \textbf{0.736} \\
\midrule
ramen   & SAM\cite{sam}           & 0.533 & -- & 0.747 & -- & 0.926 & -- & 0.735 & -- \\
        & GARField\cite{garfield}       & 0.792 & -- & 0.907 & -- & 0.955 & -- & 0.885 & -- \\
\cmidrule(lr){2-10}
        & SAM$^{*}$      & 0.538 & -- & 0.916 & -- & \textbf{0.944} & -- & 0.799 & -- \\
        & GARField$^{*}$ & 0.754 & 0.749 & 0.507 & 0.507 & 0.937 & \textbf{0.936} & 0.733 & 0.731 \\
        & OmniSeg3D$^{*}$& 0.484 & 0.483 & 0.813 & 0.800 & 0.190 & 0.161 & 0.496 & 0.481 \\
        & Ours           & \textbf{0.823} & \textbf{0.806} & \textbf{0.916} & \textbf{0.913} & 0.838 & 0.811 & \textbf{0.859} & \textbf{0.843} \\
\midrule
living\_room
        & SAM\cite{sam}           & 0.853 & -- & 0.742 & -- & 0.886 & -- & 0.827 & -- \\
        & GARField\cite{garfield}      & 0.905 & -- & 0.807 & -- & 0.944 & -- & 0.885 & -- \\
\cmidrule(lr){2-10}
        & SAM$^{*}$      & 0.850 & -- & \textbf{0.727} & -- & 0.851 & -- & 0.809 & -- \\
        & GARField$^{*}$ & 0.863 & 0.863 & 0.698 & \textbf{0.690} & \textbf{0.936} & \textbf{0.935} & \textbf{0.832} & \textbf{0.829} \\
        & OmniSeg3D$^{*}$& -- & -- & -- & -- & -- & -- & -- & -- \\
        & Ours           & \textbf{0.883} & \textbf{0.876} & 0.660 & 0.656 & 0.929 & 0.928 & 0.824 & 0.820 \\
\midrule
mean    & SAM\cite{sam}           & 0.594 & -- & 0.741 & -- & 0.887 & -- & 0.741 & -- \\
        & GARField\cite{garfield}      & 0.846 & -- & 0.843 & -- & 0.933 & -- & 0.874 & -- \\
\cmidrule(lr){2-10}
        & SAM$^{*}$      & 0.557 & -- & 0.742 & -- & \textbf{0.901} & -- & 0.733 & -- \\
        & GARField$^{*}$ & 0.824 & 0.823 & 0.607 & 0.587 & 0.815 & 0.784 & 0.749 & 0.731 \\
        & Ours           & \textbf{0.848} & \textbf{0.840} & \textbf{0.813} & \textbf{0.809} & 0.844 & \textbf{0.828} & \textbf{0.835} & \textbf{0.826} \\
\bottomrule
\end{tabular}
}
\end{table}

\subsection{Group Recall Protocol}
\label{app:group_recall_protocol}
For GARField$^{*}$, we keep its official evaluation loop, which renders scale-conditioned features at 41 scales decreasing from 2.0 to 0.0 and, at each scale, splits every current leaf with HDBSCAN at a fixed cluster-selection epsilon of 0.05, with the other settings as described below. For H2G group recall, we render the evaluation view once and project the rendered Lorentz features to the tangent space at the origin. Since H2G has a single feature field, re-clustering it at a fixed epsilon would regenerate the same groups, so we decrease the cluster-selection epsilon instead of the scale. Candidate groups are then generated by recursively splitting the current leaf regions with HDBSCAN\cite{hdbscan} in tangent space. We sweep the HDBSCAN cluster-selection epsilon from 0.50 down to 0.002 using a coarse-to-fine schedule: 0.05 steps from 0.50 to 0.20, 0.025 steps from 0.20 to 0.10, 0.01--0.015 steps from 0.10 to 0.04, 0.005 steps from 0.04 to 0.02, and smaller steps down to 0.002.
At each epsilon, every current leaf with at least 30 pixels is clustered with HDBSCAN using \texttt{min\_samples}=30 and \texttt{min\_cluster\_size}=30. Noise pixels are assigned to the nearest non-noise cluster, and clusters with fewer than 30 pixels are discarded. Each accepted cluster is added as a candidate group and becomes a new leaf for subsequent, smaller-epsilon splits. Candidate generation is capped at 10,000 nodes, although this cap is not reached in our reported results.

For native group recall, all generated candidates are used. For budgeted recall, we evaluate proposal budgets \(K\in\{50,100,150,180,300,500,750,1000,1500,2000\}\). When a scene has fewer than \(K\) H2G candidates, all available candidates are used. Random budget sampling is repeated for 100 seeds with seed 0 as the base seed. For each ground-truth group, we report the best IoU over the selected candidate set, and compute mIoU, \(R@0.50\), and \(R@0.75\). The reported AUC values are normalized areas under the corresponding budget curves.

\begin{table*}[!t]
\centering
\caption{Scene-wise group recall on the GARField group annotation benchmark. Budgeted rows use random proposal sampling averaged over 100 seeds. AUC is computed from the random budget curve with an origin point at zero budget.}
\label{tab:group_recall_scene_results}
{\tiny
\setlength{\tabcolsep}{2.8pt}
\renewcommand{\arraystretch}{0.92}
\begin{tabular}{llcccccc}
\toprule
& & \multicolumn{3}{c}{GARField$^{*}$} & \multicolumn{3}{c}{Ours} \\
\cmidrule(lr){3-5}\cmidrule(lr){6-8}
Scene & Budget & mIoU & \(R@0.50\) & \(R@0.75\) & mIoU & \(R@0.50\) & \(R@0.75\) \\
\midrule
bouquet & Candidates & \multicolumn{3}{c}{2220} & \multicolumn{3}{c}{310} \\
 & Native & 0.717 & 0.929 & 0.429 & 0.719 & 0.786 & 0.643 \\
 & \(K=50\) & 0.217 & 0.185 & 0.069 & 0.253 & 0.214 & 0.111 \\
 & \(K=100\) & 0.352 & 0.341 & 0.139 & 0.419 & 0.391 & 0.231 \\
 & \(K=150\) & 0.430 & 0.440 & 0.176 & 0.533 & 0.533 & 0.349 \\
 & \(K=180\) & 0.463 & 0.483 & 0.202 & 0.558 & 0.556 & 0.366 \\
 & \(K=300\) & 0.588 & 0.683 & 0.280 & 0.710 & 0.772 & 0.626 \\
 & \(K=500\) & 0.650 & 0.792 & 0.347 & 0.719 & 0.786 & 0.643 \\
 & \(K=750\) & 0.688 & 0.878 & 0.402 & 0.719 & 0.786 & 0.643 \\
 & \(K=1000\) & 0.699 & 0.902 & 0.411 & 0.719 & 0.786 & 0.643 \\
 & \(K=1500\) & 0.712 & 0.927 & 0.424 & 0.719 & 0.786 & 0.643 \\
 & \(K=2000\) & 0.716 & 0.929 & 0.429 & 0.719 & 0.786 & 0.643 \\
 & \(\mathrm{AUC}_{500}\) & 0.479 & 0.537 & 0.224 & 0.564 & 0.591 & 0.446 \\
 & \(\mathrm{AUC}_{2000}\) & 0.645 & 0.811 & 0.365 & 0.680 & 0.737 & 0.594 \\
\midrule
living\_room & Candidates & \multicolumn{3}{c}{3256} & \multicolumn{3}{c}{342} \\
 & Native & 0.724 & 0.778 & 0.667 & 0.722 & 0.889 & 0.667 \\
 & \(K=50\) & 0.067 & 0.057 & 0.038 & 0.186 & 0.186 & 0.118 \\
 & \(K=100\) & 0.125 & 0.117 & 0.077 & 0.329 & 0.361 & 0.234 \\
 & \(K=150\) & 0.190 & 0.172 & 0.120 & 0.419 & 0.481 & 0.311 \\
 & \(K=180\) & 0.211 & 0.197 & 0.130 & 0.487 & 0.571 & 0.383 \\
 & \(K=300\) & 0.323 & 0.309 & 0.224 & 0.662 & 0.800 & 0.596 \\
 & \(K=500\) & 0.425 & 0.417 & 0.320 & 0.722 & 0.889 & 0.667 \\
 & \(K=750\) & 0.521 & 0.538 & 0.449 & 0.722 & 0.889 & 0.667 \\
 & \(K=1000\) & 0.582 & 0.621 & 0.509 & 0.722 & 0.889 & 0.667 \\
 & \(K=1500\) & 0.652 & 0.706 & 0.606 & 0.722 & 0.889 & 0.667 \\
 & \(K=2000\) & 0.679 & 0.743 & 0.638 & 0.722 & 0.889 & 0.667 \\
 & \(\mathrm{AUC}_{500}\) & 0.254 & 0.243 & 0.176 & 0.514 & 0.613 & 0.442 \\
 & \(\mathrm{AUC}_{2000}\) & 0.512 & 0.540 & 0.447 & 0.670 & 0.820 & 0.610 \\
\midrule
ramen & Candidates & \multicolumn{3}{c}{1438} & \multicolumn{3}{c}{366} \\
 & Native & 0.773 & 0.864 & 0.682 & 0.813 & 0.909 & 0.773 \\
 & \(K=50\) & 0.255 & 0.223 & 0.145 & 0.270 & 0.237 & 0.160 \\
 & \(K=100\) & 0.358 & 0.329 & 0.222 & 0.430 & 0.405 & 0.269 \\
 & \(K=150\) & 0.433 & 0.400 & 0.286 & 0.538 & 0.537 & 0.376 \\
 & \(K=180\) & 0.478 & 0.448 & 0.330 & 0.605 & 0.616 & 0.447 \\
 & \(K=300\) & 0.549 & 0.543 & 0.399 & 0.757 & 0.830 & 0.666 \\
 & \(K=500\) & 0.638 & 0.651 & 0.505 & 0.813 & 0.909 & 0.773 \\
 & \(K=750\) & 0.707 & 0.750 & 0.587 & 0.813 & 0.909 & 0.773 \\
 & \(K=1000\) & 0.745 & 0.802 & 0.631 & 0.813 & 0.909 & 0.773 \\
 & \(K=1500\) & 0.773 & 0.864 & 0.682 & 0.813 & 0.909 & 0.773 \\
 & \(K=2000\) & 0.773 & 0.864 & 0.682 & 0.813 & 0.909 & 0.773 \\
 & \(\mathrm{AUC}_{500}\) & 0.471 & 0.458 & 0.338 & 0.609 & 0.647 & 0.508 \\
 & \(\mathrm{AUC}_{2000}\) & 0.675 & 0.723 & 0.563 & 0.762 & 0.844 & 0.706 \\
\midrule
teatime & Candidates & \multicolumn{3}{c}{1129} & \multicolumn{3}{c}{241} \\
 & Native & 0.819 & 1.000 & 0.700 & 0.869 & 1.000 & 0.900 \\
 & \(K=50\) & 0.203 & 0.194 & 0.105 & 0.315 & 0.328 & 0.242 \\
 & \(K=100\) & 0.357 & 0.377 & 0.212 & 0.519 & 0.576 & 0.434 \\
 & \(K=150\) & 0.419 & 0.481 & 0.258 & 0.673 & 0.769 & 0.614 \\
 & \(K=180\) & 0.489 & 0.560 & 0.319 & 0.733 & 0.836 & 0.698 \\
 & \(K=300\) & 0.610 & 0.720 & 0.445 & 0.869 & 1.000 & 0.900 \\
 & \(K=500\) & 0.738 & 0.907 & 0.582 & 0.869 & 1.000 & 0.900 \\
 & \(K=750\) & 0.794 & 0.977 & 0.662 & 0.869 & 1.000 & 0.900 \\
 & \(K=1000\) & 0.816 & 0.998 & 0.699 & 0.869 & 1.000 & 0.900 \\
 & \(K=1500\) & 0.819 & 1.000 & 0.700 & 0.869 & 1.000 & 0.900 \\
 & \(K=2000\) & 0.819 & 1.000 & 0.700 & 0.869 & 1.000 & 0.900 \\
 & \(\mathrm{AUC}_{500}\) & 0.506 & 0.591 & 0.359 & 0.699 & 0.797 & 0.689 \\
 & \(\mathrm{AUC}_{2000}\) & 0.732 & 0.889 & 0.602 & 0.826 & 0.949 & 0.847 \\
\bottomrule
\end{tabular}
}
\end{table*}

\subsection{Full 3D Completeness Results}
\label{app:full_completeness}
Table~\ref{tab:completeness_scene_results} reports the scene-wise 3D Completeness results summarized in Table~\ref{tab:completeness_results}, together with OmniSeg3D$^{*}$.

\subsection{Full Group Recall Results}
\label{app:full_group_recall}
Table~\ref{tab:group_recall_scene_results} reports the scene-wise group recall results summarized in Table~\ref{tab:group_recall_budget}, including the budgets omitted there.

\section{Hyperbolic versus Euclidean geometry}
\label{app:geometry}

\paragraph{Euclidean counterpart}
The output of the grouping head is mapped onto the unit hypersphere \(\mathbb S^D\subset\mathbb R^{D+1}\) by the exponential map at a fixed pole. The leaf and root angular terms use the cosine distance, prototypes and root centroids use the arithmetic mean re-projected onto the sphere, compactness uses the Euclidean distance, and the LCA surrogate uses the Euclidean norm of the point on the prototype segment closest to the origin. The loss weights, temperatures, RGB checkpoint, learning-rate schedule (\(10^{-3}\) to \(10^{-4}\) over 20k iterations), and seed are those of the main runs. Evaluation follows the same protocols, with the logarithmic map replaced by the identity.

\paragraph{Level-wise 3D Completeness}
Table~\ref{tab:geo_level} reports the same comparison per level under view-wise (V) and level-wise (L) threshold selection. Lorentz features score higher than Euclidean features at every level for both dimensions and both threshold settings, and the 16-dimensional Lorentz field exceeds the 256-dimensional Euclidean field in mean under both settings.

\begin{table}[!b]
\centering
\caption{3D Completeness of unit-norm Euclidean and Lorentz features under view-wise (V) and level-wise (L) threshold selection.}
\label{tab:geo_level}
{
\small
\setlength{\tabcolsep}{4.5pt}
\begin{tabular}{lccccccccc}
\toprule
& & \multicolumn{2}{c}{Fine} & \multicolumn{2}{c}{Medium} & \multicolumn{2}{c}{Coarse} & \multicolumn{2}{c}{Mean} \\
\cmidrule(lr){3-4}\cmidrule(lr){5-6}\cmidrule(lr){7-8}\cmidrule(lr){9-10}
Geometry & \(D\) & V & L & V & L & V & L & V & L \\
\midrule
Euclidean & 16  & 0.726 & 0.712 & 0.774 & 0.729 & 0.703 & 0.613 & 0.734 & 0.685 \\
Lorentz   & 16  & \textbf{0.854} & \textbf{0.783} & \textbf{0.806} & \textbf{0.801} & \textbf{0.791} & \textbf{0.758} & \textbf{0.817} & \textbf{0.780} \\
\midrule
Euclidean & 256 & 0.752 & 0.742 & 0.755 & 0.739 & 0.775 & 0.721 & 0.761 & 0.734 \\
Lorentz   & 256 & \textbf{0.848} & \textbf{0.840} & \textbf{0.813} & \textbf{0.809} & \textbf{0.844} & \textbf{0.828} & \textbf{0.835} & \textbf{0.826} \\
\bottomrule
\end{tabular}
}
\end{table}
\begin{table}[!b]
\centering
\caption{CTR at \(D=256\).}
\label{tab:geo_ctr}
{
\small
\begin{tabular}{lcccc}
\toprule
Geometry & Fine & Medium & Coarse & Mean \\
\midrule
Euclidean & 0.988 & 0.979 & 0.918 & 0.962 \\
Lorentz   & \textbf{0.990} & \textbf{0.995} & \textbf{0.978} & \textbf{0.988} \\
\bottomrule
\end{tabular}
}
\end{table}

\begin{table}[!b]
\centering
\caption{Native group recall at \(D=256\), Euclidean / Lorentz. Lorentz values come from a separate run of the recursive clustering and differ slightly from Table~\ref{tab:group_recall_scene_results}.}
\label{tab:geo_recall}
{
\small
\setlength{\tabcolsep}{4pt}
\begin{tabular}{lcccc}
\toprule
Scene & Candidates & mIoU & \(R@0.50\) & \(R@0.75\) \\
\midrule
bouquet      & 95 / 305   & 0.631 / \textbf{0.719} & 0.714 / \textbf{0.786} & 0.571 / \textbf{0.643} \\
teatime      & 73 / 248   & 0.595 / \textbf{0.868} & 0.600 / \textbf{1.000} & 0.600 / \textbf{0.900} \\
ramen        & 84 / 367   & 0.724 / \textbf{0.813} & 0.727 / \textbf{0.909} & 0.545 / \textbf{0.773} \\
living\_room & 209 / 339  & 0.518 / \textbf{0.722} & 0.556 / \textbf{0.889} & 0.222 / \textbf{0.667} \\
\midrule
Mean                  & 115.2 / 314.8 & 0.617 / \textbf{0.781} & 0.649 / \textbf{0.896} & 0.485 / \textbf{0.746} \\
\bottomrule
\end{tabular}
}
\end{table}

\paragraph{CTR and group recall}
3D Completeness selects the best of 1000 thresholds for each view, which can hide level boundaries that fall at different affinities across views. Tables~\ref{tab:geo_ctr} and~\ref{tab:geo_recall} report measurements at \(D=256\) that involve no per-view selection. With one threshold per level shared across views, Euclidean features retain less of their view-wise score than Lorentz features, most clearly at the coarse level, so Euclidean level boundaries are not consistent across views. For group recall, both geometries use the recursive HDBSCAN of Appendix~\ref{app:group_recall_protocol} with identical hyperparameters. The Euclidean field yields a more compact candidate pool, but Lorentz gives higher native mIoU, \(R@0.50\), and \(R@0.75\) on every scene, including \texttt{bouquet} and \texttt{teatime}, where 3D Completeness shows no difference. A smaller pool with lower recall suggests that Euclidean features merge groups that should be separated, so the recursive clustering stops splitting early.

\section{LCA Ordering Accuracy}
\label{app:lca_acc}

In the ablation of Table~\ref{tab:ablation}, removing the LCA ordering loss changes 3D Completeness the least among the four components. 3D Completeness, however, does not measure ordering, so we evaluate the merge order of the learned field directly.

\paragraph{Depth of the supervision trees}
LCA ordering constrains the relative merge order among leaf prototypes and is most informative when many distinct ancestor depths must be ordered. With the flattening threshold used in our experiments, every supervision tree has depth at most two (Appendix~\ref{app:hierarchy_examples}). The lowest common ancestor of two distinct leaves can therefore lie at only two depths, 1 for siblings and 0 for cousins, so exactly one type of ordering relation exists. The evaluation annotations likewise provide only two or three levels per scene, which leaves few orderings to get right or wrong.

\paragraph{Triplet accuracy}
We measure whether the learned field preserves merge order as a triplet accuracy. If \(i\) and \(j\) join at a deeper ancestor than \(i\) and \(k\) in the supervision tree, the learned hierarchy score should satisfy \(s(i,j)>s(i,k)\):
$$
\mathrm{LCA\text{-}Acc}
=
\frac{1}{|\mathcal Q|}
\sum_{(i,j,k)\in\mathcal Q}
\mathbb{I}\left[s(i,j)>s(i,k)\right],
$$
where \(\mathbb{I}[\cdot]\) is the indicator function and \(\mathcal Q\) is the set of evaluation triplets. Here \(s(a,b)\) is the LCA-depth surrogate \(d_o\) of Section~\ref{sec:method}, computed between rendered features of pixels sampled from leaves \(a\) and \(b\); the Euclidean field uses the Euclidean norm instead, as in Appendix~\ref{app:geometry}. Triplets are drawn from the supervision tree of each view and pooled over the views of a scene; an uninformative predictor scores 0.5.

\begin{table}[h]
\centering
\caption{LCA-Acc of the learned grouping fields. Euclidean is the unit-norm counterpart of Appendix~\ref{app:geometry} trained with all four losses. Chance is 0.5.}
\label{tab:lca_acc}
{
\small
\setlength{\tabcolsep}{4.5pt}
\begin{tabular}{lccccc}
\toprule
Model & teatime & bouquet & ramen & living\_room & Mean \\
\midrule
Euclidean                                         & \textbf{0.9737} & 0.8171 & 0.9377 & 0.9345 & 0.9158 \\
Lorentz w/o \(\mathcal L_{\mathrm{LCA}}\)          & 0.9700 & 0.8325 & 0.9452 & 0.9395 & 0.9218 \\
Lorentz (full)                                    & \textbf{0.9737} & \textbf{0.8545} & \textbf{0.9526} & \textbf{0.9505} & \textbf{0.9328} \\
\bottomrule
\end{tabular}
}
\end{table}

\paragraph{Effect of geometry and the LCA loss}
Table~\ref{tab:lca_acc} shows that the learned field preserves the merge order of the hierarchy far above chance on every scene. Much of this preservation comes from the curvature. With the losses fixed, changing only the geometry raises the mean more than removing \(\mathcal L_{\mathrm{LCA}}\) lowers it with the geometry fixed, and the Lorentz field without \(\mathcal L_{\mathrm{LCA}}\) still scores above the Euclidean field with it in mean. The contribution of \(\mathcal L_{\mathrm{LCA}}\) is small but consistent. It reduces ordering errors, \(1-\mathrm{LCA\text{-}Acc}\), by 14\% on average and improves every scene; the absolute gain varies across scenes, but the relative error reduction stays within 12--18\%.

\paragraph{Fidelity to the supervision trees}
LCA-Acc measures fidelity to our own supervision trees, not whether that order is correct, so a field that follows a systematically biased tree would also score well. Measuring correctness requires an external, human-annotated hierarchy deeper than the two or three levels available to us, which remains open. The comparison itself is nonetheless sound, since all three models are evaluated against the same trees. Because per-view trees do not agree with one another, the ceiling of this metric is below 1.0. A 3D field can express only one view-independent ordering and cannot follow the idiosyncrasies of individual views, so high average agreement indicates an ordering close to the consensus of the views rather than one memorized from any of them.

\section{Hyperparameter Sensitivity}
\label{app:sensitivity}

We vary the loss weights and the curvature around the main configuration. Each run retrains only the grouping field from the same RGB checkpoint, with everything else (other weights, geometry, dimension, temperatures, margins, and seed) held at the main run.

\paragraph{Loss weights}
Each of the four hierarchy weights and \(\lambda_{\mathrm{norm}}\) is scaled to \(0.5\times\) and \(2\times\) its default (Table~\ref{tab:sens_weights}). Across this fourfold range, the mean changes only slightly for every hierarchy weight, and the default is the best of these nine settings. \(\lambda_{\mathrm{norm}}\) moves the mean even less, and both of its settings stay within the run-to-run variation of Appendix~\ref{app:reproducibility}. The field is therefore robust to all five weights.

\begin{table}[h]
\centering
\caption{3D Completeness (view-wise) when each loss weight is scaled to \(0.5\times\) or \(2\times\) its default.}
\label{tab:sens_weights}
{
\small
\begin{tabular}{llcccc}
\toprule
Weight & Factor & Fine & Medium & Coarse & Mean \\
\midrule
--- & \(1.0\times\) (main) & 0.848 & 0.813 & 0.844 & 0.835 \\
\midrule
\(\lambda_{\mathrm{leaf}}\) & \(0.5\times\) & 0.837 & 0.802 & 0.822 & 0.820 \\
                            & \(2\times\)   & 0.863 & 0.792 & 0.786 & 0.814 \\
\(\lambda_{\mathrm{root}}\) & \(0.5\times\) & 0.860 & 0.813 & 0.810 & 0.828 \\
                            & \(2\times\)   & 0.836 & 0.795 & 0.817 & 0.816 \\
\(\lambda_{\mathrm{comp}}\) & \(0.5\times\) & 0.849 & 0.795 & 0.819 & 0.821 \\
                            & \(2\times\)   & 0.846 & 0.808 & 0.825 & 0.826 \\
\(\lambda_{\mathrm{LCA}}\)  & \(0.5\times\) & 0.848 & 0.804 & 0.806 & 0.819 \\
                            & \(2\times\)   & 0.849 & 0.806 & 0.796 & 0.817 \\
\(\lambda_{\mathrm{norm}}\) & \(0.5\times\) & 0.848 & 0.827 & 0.842 & 0.839 \\
                            & \(2\times\)   & 0.846 & 0.825 & 0.838 & 0.836 \\
\bottomrule
\end{tabular}
}
\end{table}

\paragraph{Curvature}
Table~\ref{tab:sens_curvature} varies the curvature \(c\) over two orders of magnitude. \(c=1.0\) is best in mean but not on every scene: \texttt{teatime} prefers \(c=2\) and \texttt{ramen} prefers \(c=10\), while \texttt{bouquet} and \texttt{living\_room} are best at \(c=1.0\). Curvature is thus worth tuning, and we have not optimized it. All loss weights and the curvature are fixed across the four scenes; none is tuned per scene.

\begin{table}[h]
\centering
\caption{3D Completeness (view-wise) for different curvatures \(c\). Scene columns are scene means, level columns average each level over the scenes that annotate it, and Mean follows Appendix~\ref{app:aggregation}.}
\label{tab:sens_curvature}
{
\small
\setlength{\tabcolsep}{4pt}
\begin{tabular}{lcccccccc}
\toprule
& \multicolumn{4}{c}{Scene} & \multicolumn{3}{c}{Level} & \\
\cmidrule(lr){2-5}\cmidrule(lr){6-8}
\(c\) & teatime & bouquet & ramen & living\_room & Fine & Medium & Coarse & Mean \\
\midrule
0.1          & 0.915 & 0.662 & 0.802 & 0.734 & 0.802 & 0.740 & 0.749 & 0.764 \\
1.0 (main)   & 0.952 & \textbf{0.750} & 0.859 & \textbf{0.824} & \textbf{0.848} & \textbf{0.813} & \textbf{0.844} & \textbf{0.835} \\
2            & \textbf{0.955} & 0.709 & 0.837 & 0.786 & 0.843 & 0.772 & 0.805 & 0.807 \\
10           & 0.939 & 0.619 & \textbf{0.884} & 0.798 & 0.847 & 0.759 & 0.779 & 0.795 \\
\bottomrule
\end{tabular}
}
\end{table}

\section{Inference Cost}
\label{app:inference}

\begin{table}[!b]
\centering
\caption{Inference time per evaluation view in seconds, measured on an RTX 4090. \emph{Renders reused} shares the 41 scale-conditioned renders across hierarchy levels, and \emph{as scripted} follows the released GARField evaluation script.}
\label{tab:inference_cost}
{
\small
\begin{tabular}{lcccc}
\toprule
& & & \multicolumn{2}{c}{GARField$^{*}$} \\
\cmidrule(lr){4-5}
Scene & Levels & H2G & Renders reused & As scripted \\
\midrule
teatime          & 2 & 0.456 & 14.34 & 28.67 \\
bouquet          & 3 & 0.489 & 14.12 & 42.35 \\
ramen            & 3 & 0.488 & 14.18 & 42.54 \\
living\_room     & 3 & 0.486 & 14.17 & 42.52 \\
\midrule
Mean             &   & 0.480 & 14.20 & 39.02 \\
\bottomrule
\end{tabular}
}
\end{table}

We time inference under the 3D Completeness protocol for the four scenes and their three evaluation views, with \(369\times497\) rays per view. Only the grouping branch is timed, and the RGB head is excluded for both methods. All timings are measured on a single NVIDIA RTX 4090 with CUDA events, and each entry is the median of five timed repetitions after two warm-up runs (Table~\ref{tab:inference_cost}).

GARField cannot know the right scale for a query in advance, so its evaluation sweeps 41 scales from 2.0 to 0.0 and renders the scale-conditioned affinity field at each. Its released script repeats this sweep at every hierarchy level, which amounts to 123 renders per view for a three-level scene. We also time a variant modified in GARField's favor that renders the 41 scales once and shares them across levels. H2G has no scale axis: it renders the grouping field once and computes the tangent-space affinity map and the per-level threshold sweep from that single render. H2G is therefore more than an order of magnitude faster than either GARField variant. Its cost does not depend on the number of query scales and barely changes with the number of levels, whereas the cost of the released GARField script grows in proportion to it.

\section{Spectral Bisection Analysis}
\label{app:spectral_bisection}
Dasgupta's objective motivates a recursive top-down construction in which each branch is split by a sparsest-cut approximation. In our setting, however, each proposal group induces a dense DINO affinity graph, making exact recursive sparsest cut expensive as the number of leaves grows. We therefore use recursive spectral bisection in the main pipeline and compare it with exact sparsest cut on the same DINO graphs.

\begin{figure}[!b]
  \centering
  \includegraphics[width=\textwidth]{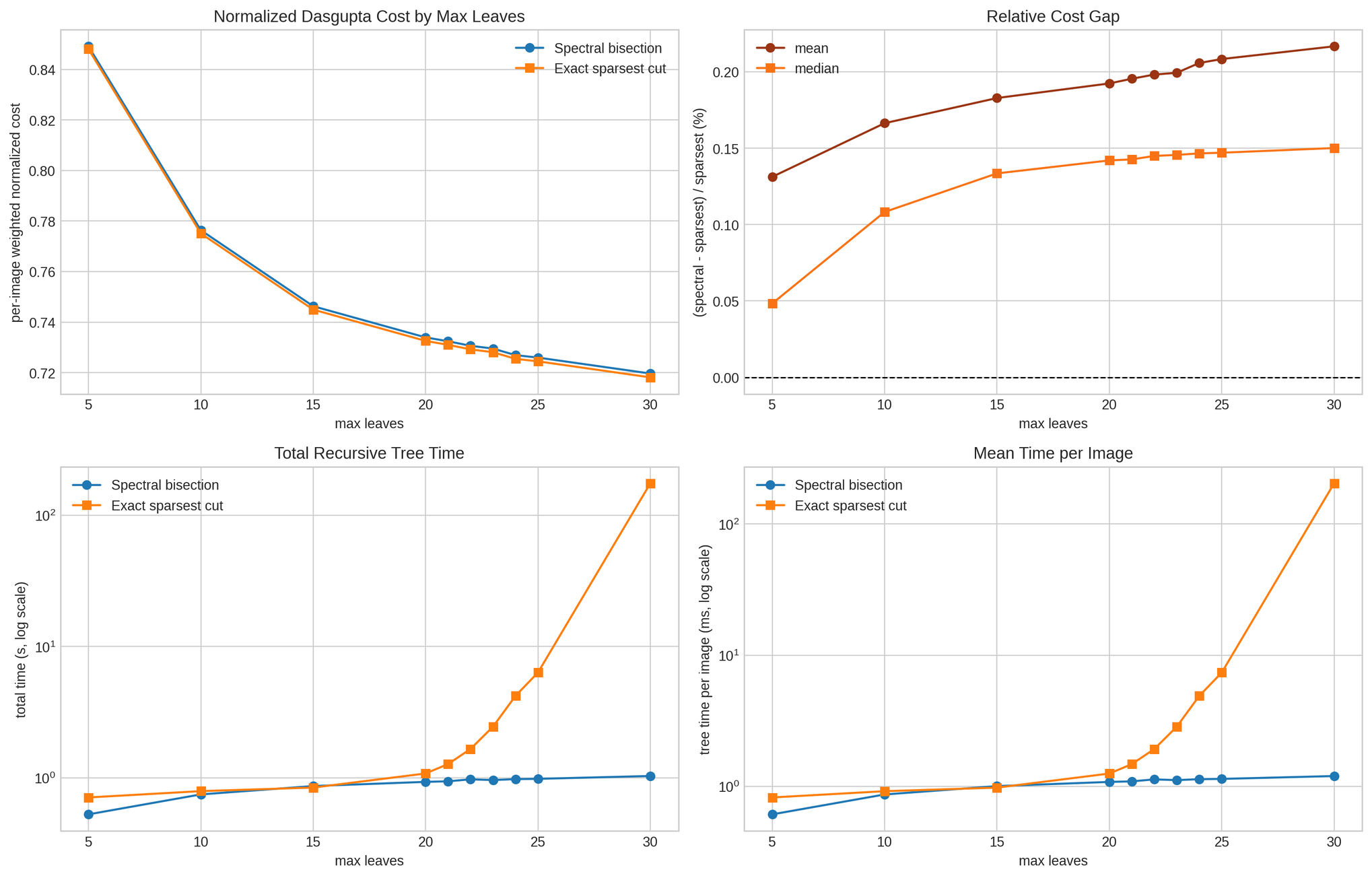}
  \caption{Comparison between recursive spectral bisection and exact recursive sparsest cut for 2D hierarchy construction. Spectral bisection closely matches the normalized Dasgupta cost of exact sparsest cut while keeping preprocessing time nearly constant. Time plots use a log scale.}
  \label{fig:dasgupta_spectral}
\end{figure}

Figure~\ref{fig:dasgupta_spectral} shows that spectral bisection closely tracks exact sparsest cut in normalized Dasgupta cost across max-leaf settings. The relative cost gap remains small: the median gap is below \(0.15\%\), and the mean gap stays below about \(0.22\%\) even at the largest setting. This indicates that the surrogate preserves the tree quality measured by Dasgupta's objective.

The computational difference is much larger. Exact recursive sparsest cut grows rapidly with the max-leaf setting, reaching over \(10^2\) seconds in total preprocessing time, whereas spectral bisection remains around one second. The same trend appears in mean per-image time, where exact sparsest cut increases by orders of magnitude while spectral bisection stays nearly flat. These results support using spectral bisection as a practical surrogate: it introduces only a minor Dasgupta-cost increase but removes the main preprocessing bottleneck for dense DINO affinity graphs.

\section{Additional Qualitative Results}
\label{app:qualitative}

\subsection{2D Hierarchy Supervision}
\label{app:hierarchy_examples}
\begin{figure}[h]
  \centering
  \includegraphics[width=\textwidth]{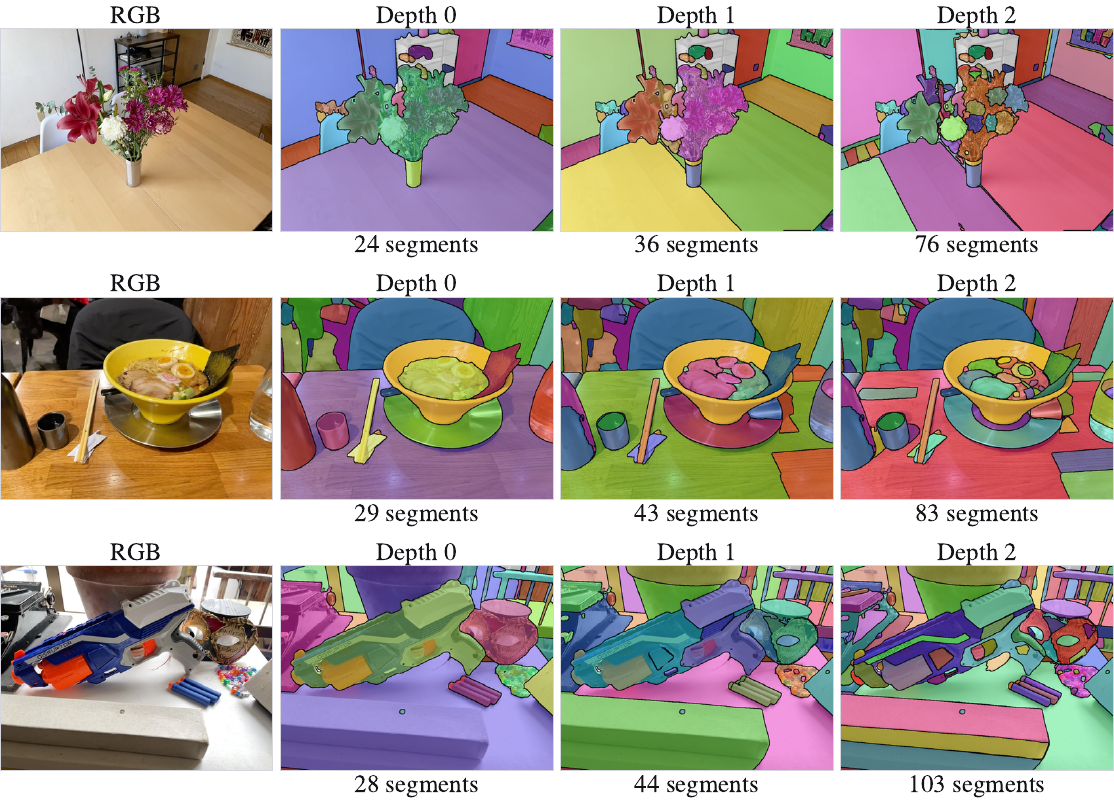}
  \caption{Additional examples of 2D hierarchy supervision, visualized as in Figure~\ref{fig:hierarchy_teatime}.}
  \label{fig:hierarchy_examples}
\end{figure}

As in Figure~\ref{fig:hierarchy_teatime}, Figure~\ref{fig:hierarchy_examples} visualizes the 2D hierarchy supervision for \texttt{bouquet}, \texttt{ramen}, and \texttt{living\_room}, with each tree of \(\mathcal T^n\) cut at depth \(d\) from its root. For example, the bowl contents in \texttt{ramen} are split into individual toppings at deeper levels. The tree depth depends on the flattening threshold. With the threshold of 0.18 used in our experiments, every tree has depth at most two, so depth 2 shows the leaf masks.

\subsection{Hyperbolic Embedding of Ground-Truth Groups}
\label{app:cosne}
\begin{figure}[h]
  \centering
  \includegraphics[width=\textwidth]{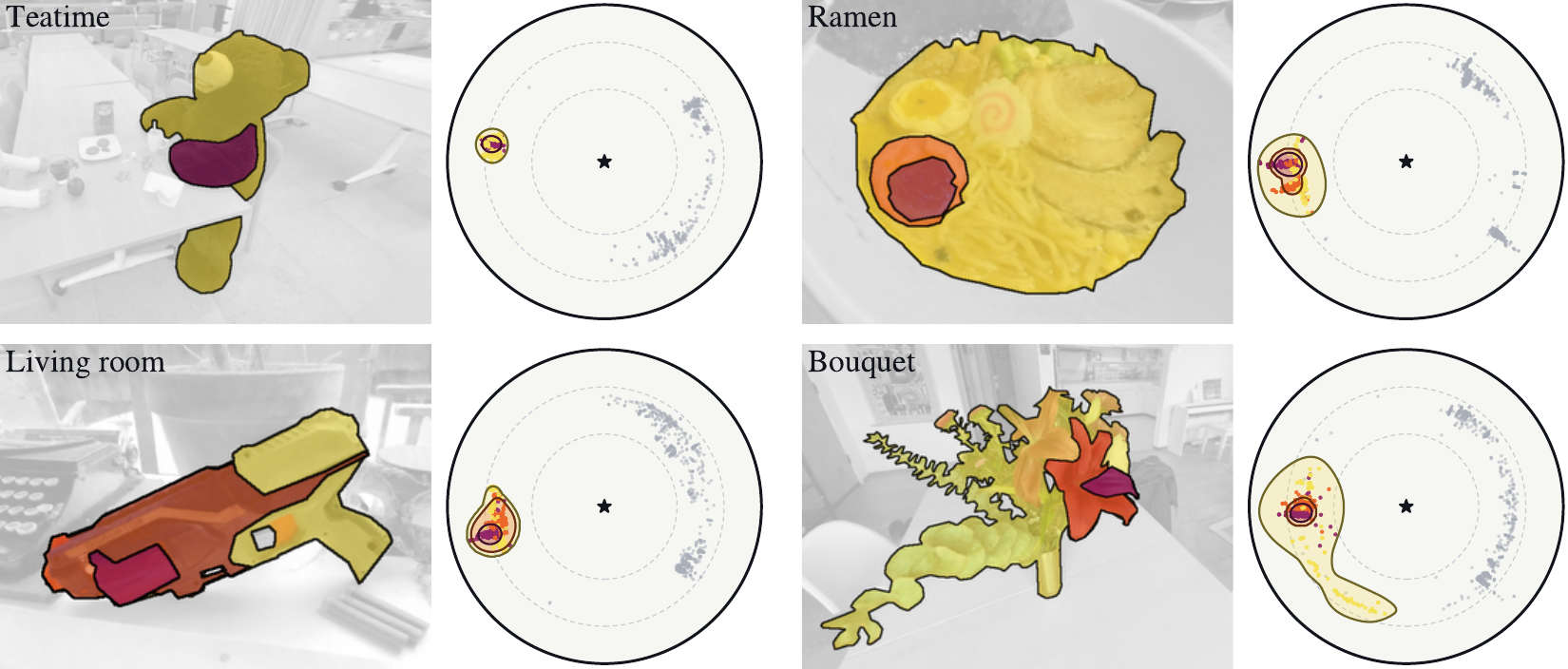}
  \caption{CO-SNE projections of H2G features. Ground-truth groups are colored from fine (purple) to coarse (yellow), gray points are context pixels, and the star marks the origin.}
  \label{fig:cosne}
\end{figure}

Figure~\ref{fig:cosne} projects rendered features of the H2G model in Table~\ref{tab:completeness_results} onto the Poincaré disk with CO-SNE\cite{cosne}, which preserves distances from the origin. For each scene, we sample 200 pixels per ground-truth group in the first evaluation view, together with as many context pixels outside the groups. The groups of each scene gather in one angular direction, apart from the context. Pixels lie at similar distances from the origin whichever group they belong to, but group midpoints move toward the origin from fine to coarse in every scene, as a coarser group spreads over a wider angle.

\subsection{K-means on the Grouping Field}
\label{app:kmeans}
\begin{figure}[htbp]
  \centering
  \includegraphics[width=\textwidth]{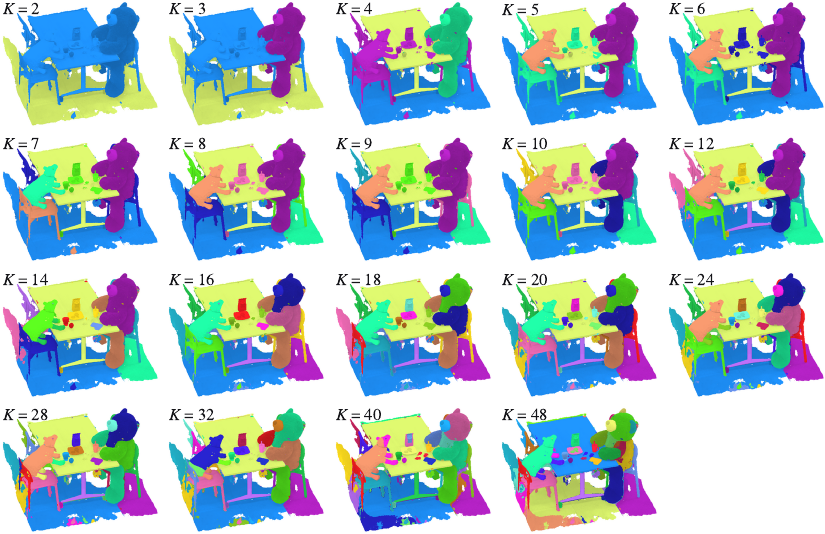}
  \caption{K-means clusters on \texttt{teatime} for \(K\) from 2 to 48.}
  \label{fig:kmeans_teatime}
\end{figure}

\begin{figure}[htbp]
  \centering
  \includegraphics[width=\textwidth]{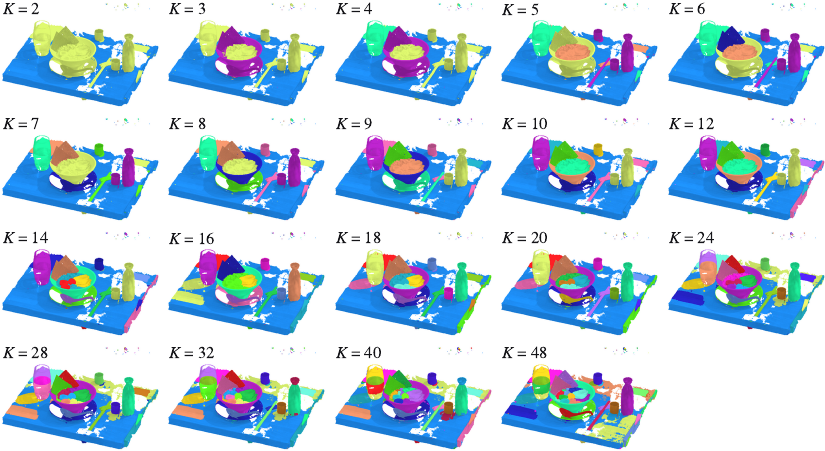}
  \caption{K-means clusters on \texttt{ramen} for \(K\) from 2 to 48.}
  \label{fig:kmeans_ramen}
\end{figure}

\begin{figure}[htbp]
  \centering
  \includegraphics[width=\textwidth]{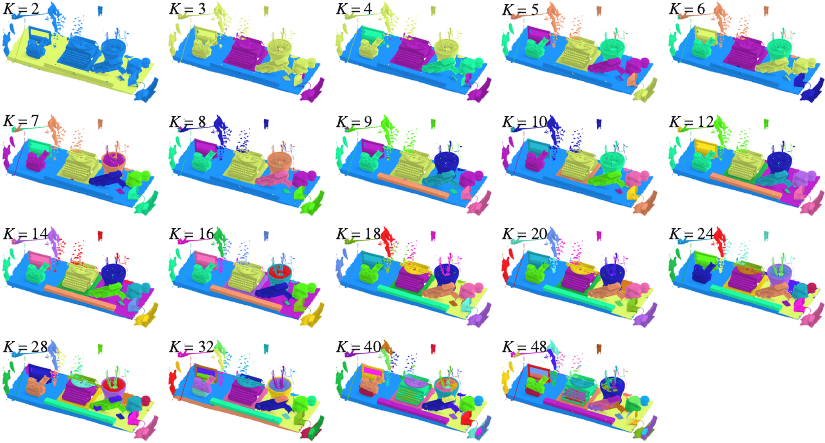}
  \caption{K-means clusters on \texttt{living\_room} for \(K\) from 2 to 48.}
  \label{fig:kmeans_living_room}
\end{figure}

\begin{figure}[htbp]
  \centering
  \includegraphics[width=\textwidth]{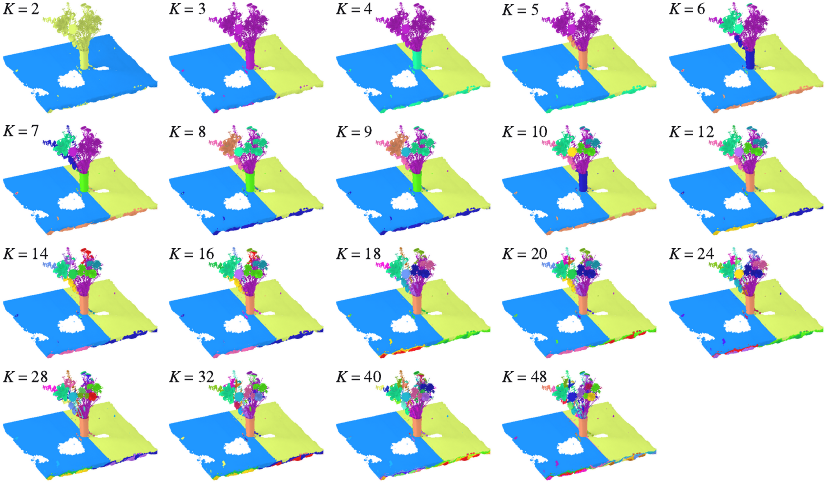}
  \caption{K-means clusters on \texttt{bouquet} for \(K\) from 2 to 48.}
  \label{fig:kmeans_bouquet}
\end{figure}

Figures~\ref{fig:kmeans_teatime}--\ref{fig:kmeans_bouquet} apply K-means to the same model for \(K\) from 2 to 48, rendering each scene cropped to a box from a novel viewpoint. As in our evaluation, the features are mapped to the tangent space at the origin, and colors are not matched across \(K\). As \(K\) grows, objects first separate from the background and then split into parts. Since K-means is not hierarchical, clusters at different \(K\) need not be nested, and large uniform surfaces can split arbitrarily.

\end{document}